\documentclass[10pt, a4paper]{article}

\usepackage[]{lrec-coling2024} 
\usepackage{multirow}
\usepackage{booktabs}
\usepackage{adjustbox}
\usepackage{listings}

\usepackage{natbib}
\usepackage{multibib}
\makeatletter
\def\@mb@citenamelist{cite,citep,citet,citealp,citealt,citepalias,citetalias}
\makeatother
\newcites{languageresource}{~}

\usepackage{graphicx}
\usepackage{tabularx}
\usepackage{soul}

\usepackage{IEEEtrantools}
\usepackage{graphicx}
\usepackage{amsmath}
\usepackage{amsthm}
\usepackage{booktabs}
\usepackage{algorithm}
\usepackage{algorithmic}
\usepackage{natbib} 
\usepackage{amssymb}
\usepackage{multirow}
\usepackage{boldline}
\usepackage{hhline}
\usepackage{amsmath}
\usepackage{xcolor}
\usepackage{pifont}
\usepackage{subfigure}
\usepackage{makecell}
\usepackage{mathrsfs}
\usepackage{utfsym}
\usepackage[normalem]{ulem}
\useunder{\uline}{\ul}{}
\usepackage[inline]{enumitem}
\usepackage{enumitem}
\usepackage{IEEEtrantools}
\usepackage{stmaryrd}
\usepackage{mathabx}

\usepackage{xcolor}
\usepackage{hyperref}
 \definecolor{darkblue}{rgb}{0, 0, 0.5}
  \hypersetup{colorlinks=true, citecolor=darkblue, linkcolor=darkblue, urlcolor=darkblue}

\usepackage{xstring}

\usepackage{color}

\title{Learning Intrinsic Dimension via Information Bottleneck for Explainable Aspect-based Sentiment Analysis}

\name{\normalsize Zhenxiao Cheng$^1$, Jie Zhou$^{1,*}$\thanks{$^*$ Corresponding author, jzhou@cs.ecnu.edu.cn.}, Wen Wu$^{1}$, Qin Chen$^1$, Liang He$^1$} 

\address{\normalsize $^1$ School of Computer Science and Technology, East China Normal University, Shanghai, China \\
}

\abstract{
Gradient-based explanation methods are increasingly used to interpret neural models in natural language processing (NLP) due to their high fidelity. Such methods determine word-level importance using dimension-level gradient values through a norm function, often presuming equal significance for all gradient dimensions. However, in the context of Aspect-based Sentiment Analysis (ABSA), our preliminary research suggests that only specific dimensions are pertinent. To address this, we propose the Information Bottleneck-based Gradient (\texttt{IBG}) explanation framework for ABSA. This framework leverages an information bottleneck to refine word embeddings into a concise intrinsic dimension, maintaining essential features and omitting unrelated information. Comprehensive tests show that our \texttt{IBG} approach considerably improves both the models' performance and interpretability by identifying sentiment-aware features.
 \\ \newline  
}

\begin{document}

\maketitleabstract

\section{Introduction}
The domain of natural language processing (NLP) has witnessed the rise of neural models that offer remarkable capabilities. Yet, the intricacies of these models often remain cloaked in layers of complexity, raising questions about their interpretability \cite{danilevsky-etal-2020-survey,ribeiro2016should,lundberg2017unified}. Gradient-based explanation methods \cite{DBLP:journals/corr/SimonyanVZ13,DBLP:journals/corr/SmilkovTKVW17,sundararajan2017axiomatic} have emerged as a prominent solution to this interpretability conundrum, offering insights into how neural models function \cite{doshi2017towards}, especially in terms of their fidelity. 

These methods pivot on the idea of ascertaining the importance of words by utilizing dimension-level gradient values, processed through a norm function. 
Formally, Gradient-based explanation methods estimate the contribution of input $x$ towards output $y$ by computing the partial derivative of $y$ w.r.t $x$.
These saliency methods can be used to enable feature importance explainability, especially on word/token-level features \cite{aubakirova2016interpreting,karlekar2018detecting}.
Then, Smooth Gradient \cite{DBLP:journals/corr/SmilkovTKVW17} and Integrated Gradients \cite{sundararajan2017axiomatic} are proposed to improve the original gradients. 
A prevalent assumption made during this process is the uniform significance attributed to every gradient dimension.


However, while this might hold true for many applications, the nuances of Aspect-based Sentiment Analysis (ABSA) present a more complex scenario. 
In our preliminary analysis of aspect-based sentiment classification tasks (Section \ref{sect:Preliminary Analysis}), we have found that this assumption is not always valid. \textbf{First}, not all dimensions are equally significant. \textbf{Second}, while the number of important dimensions varies across datasets, only a few dimensions prove essential (intrinsic dimension \cite{li2018measuring}). \textbf{Third}, the key dimensions exhibit similarity within a dataset but vary across different datasets. 
For example, dimension 401 ranks among the top 100 important dimensions for 89\% instances in the Res14 dataset. Further elaboration on these findings can be found in Section \ref{sect:Preliminary Analysis}.
This observation calls for a more discerning approach to gradient-based explanations in the ABSA context.

\begin{figure}[!t]
\begin{center}
\includegraphics[scale=0.4]{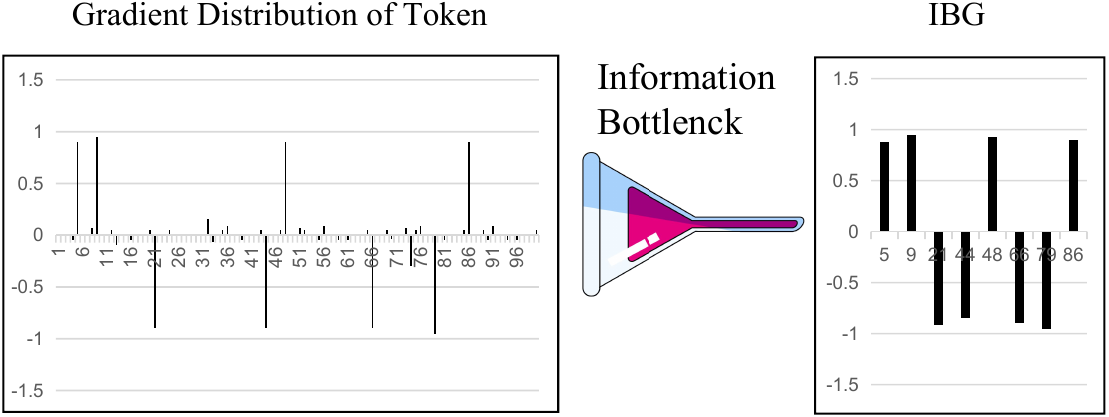} 
\caption{IBG compresses the 100 noisy dimensions of token gradient into 8 intrinsic dimensions via information bottleneck.}
\label{fig:intro}
\end{center}
\end{figure}

\begin{figure*}[!t]
\centering
\subfigure[Lap14] {\includegraphics[scale=0.235]{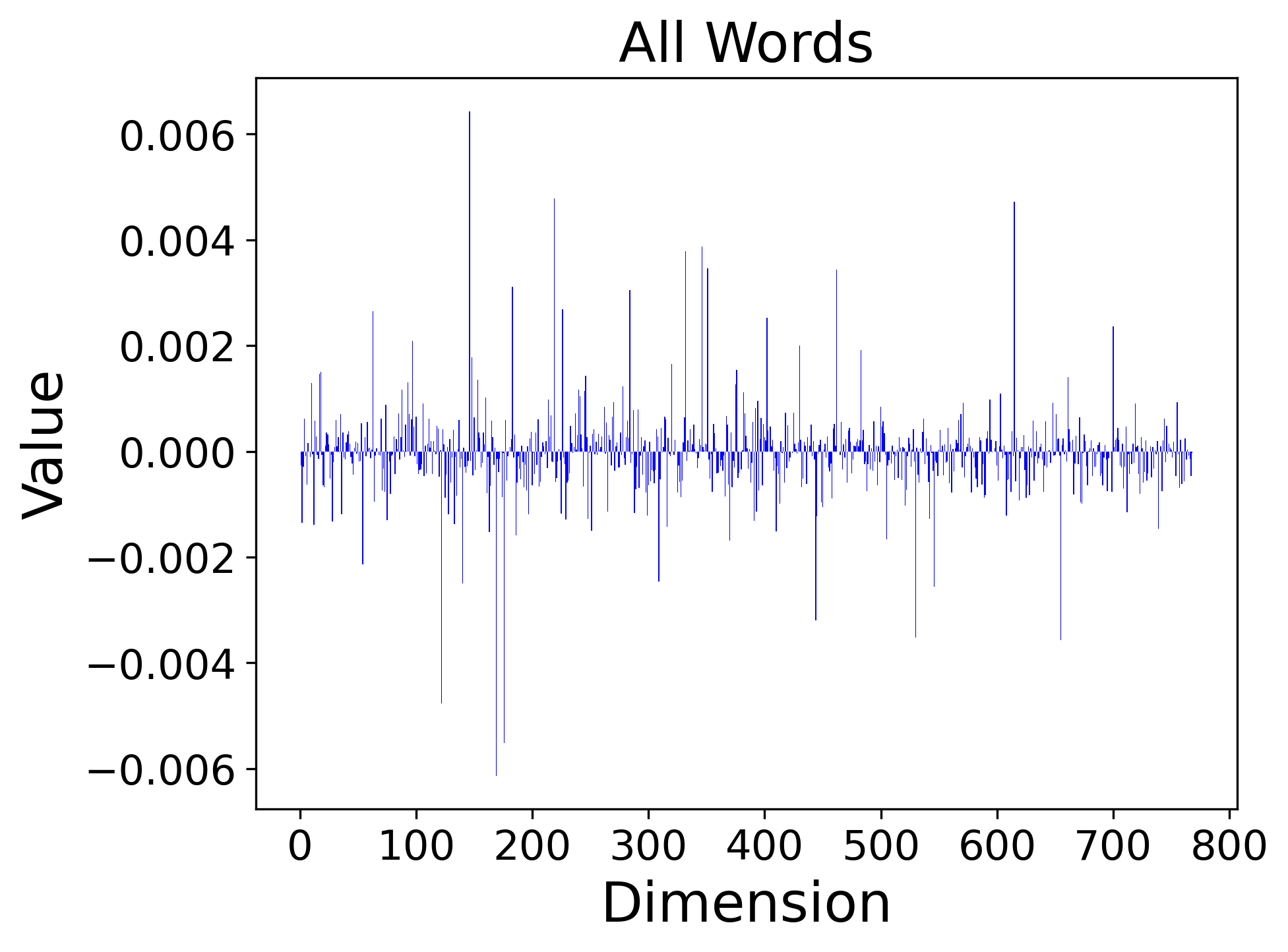}}
\subfigure[Res14] {\includegraphics[scale=0.235]{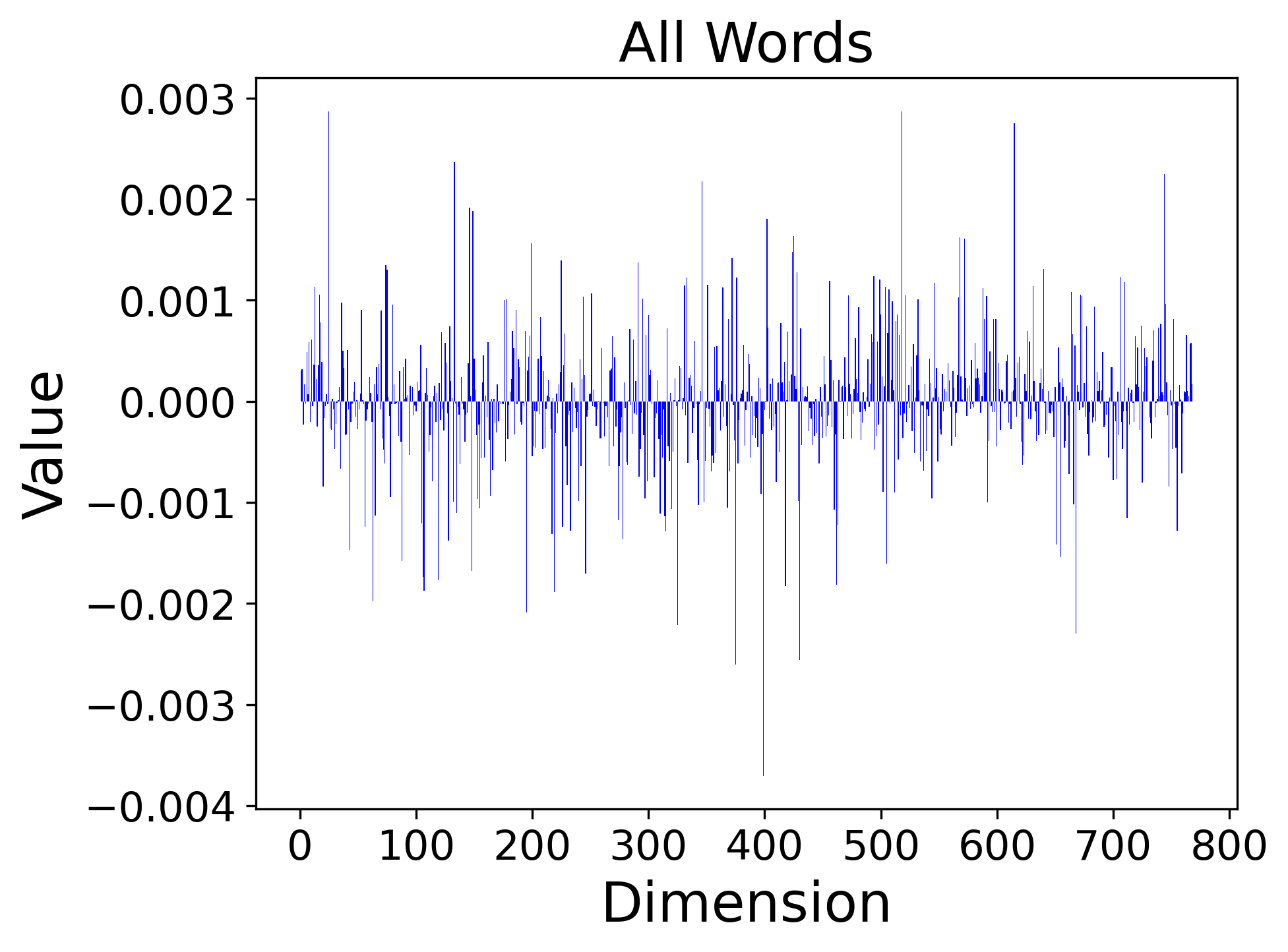}}
\subfigure[Res15] {\includegraphics[scale=0.235]{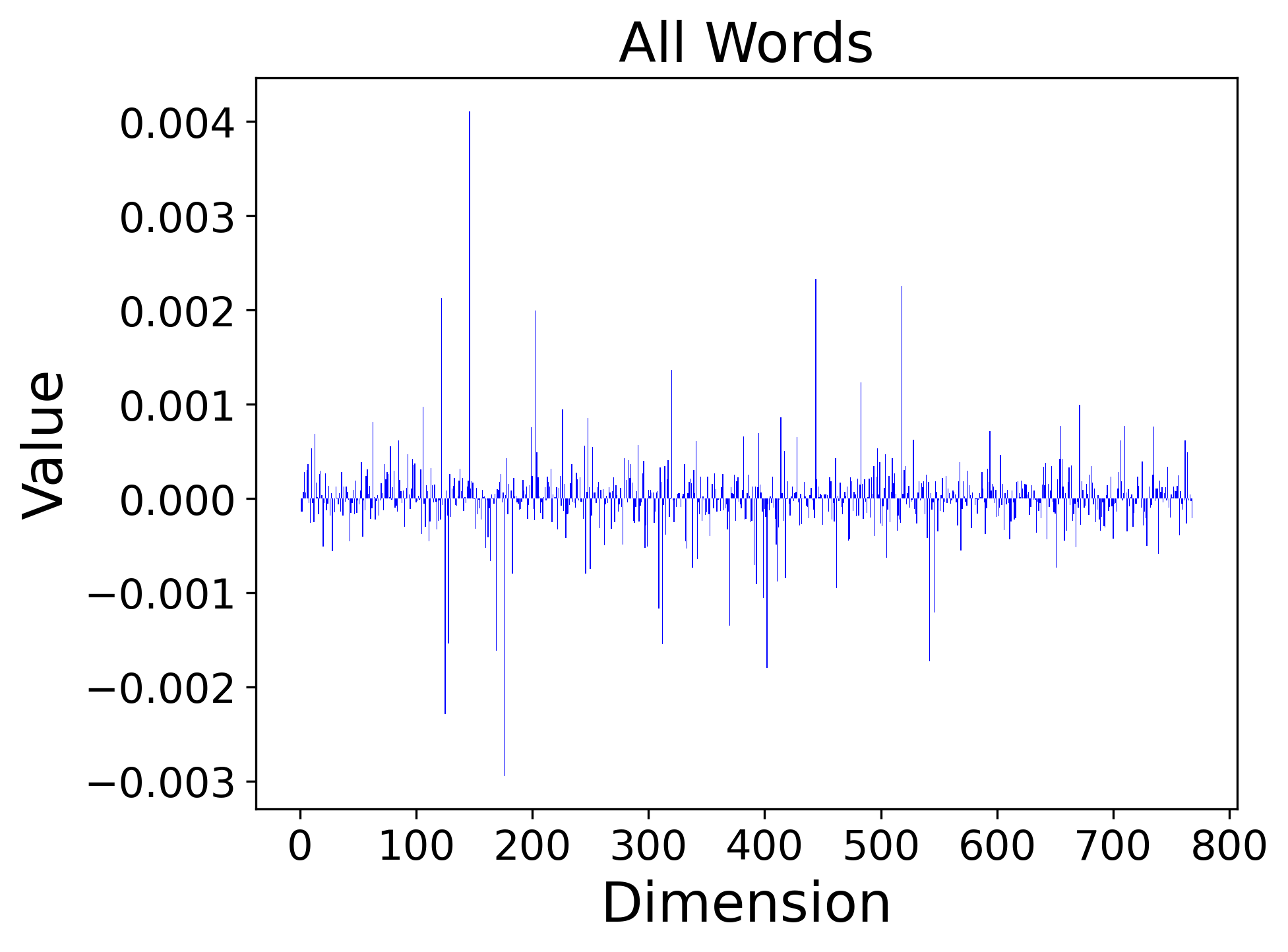}}
\subfigure[Res16] {\includegraphics[scale=0.235]{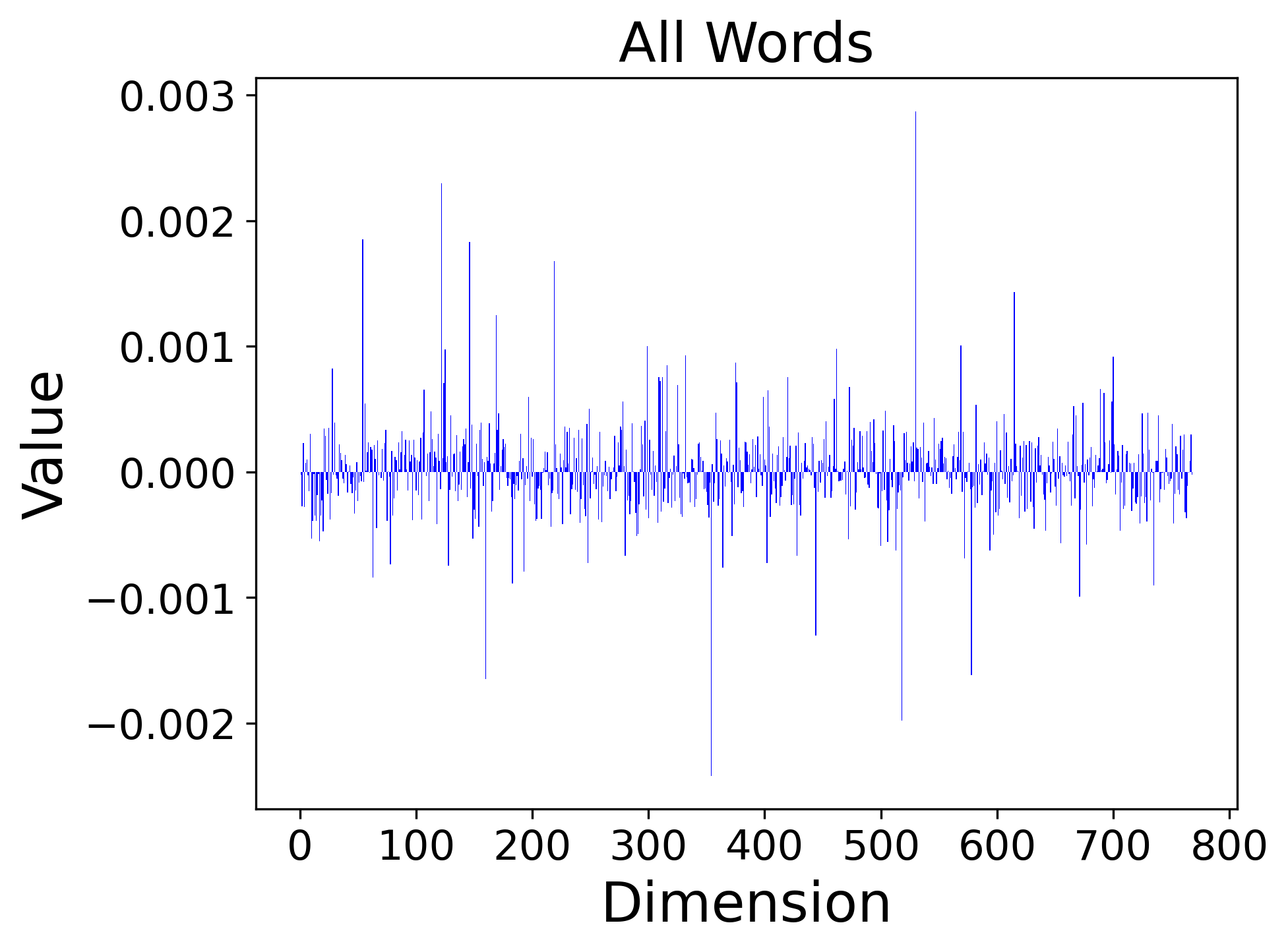}}
\subfigure[Lap14] {\includegraphics[scale=0.235]{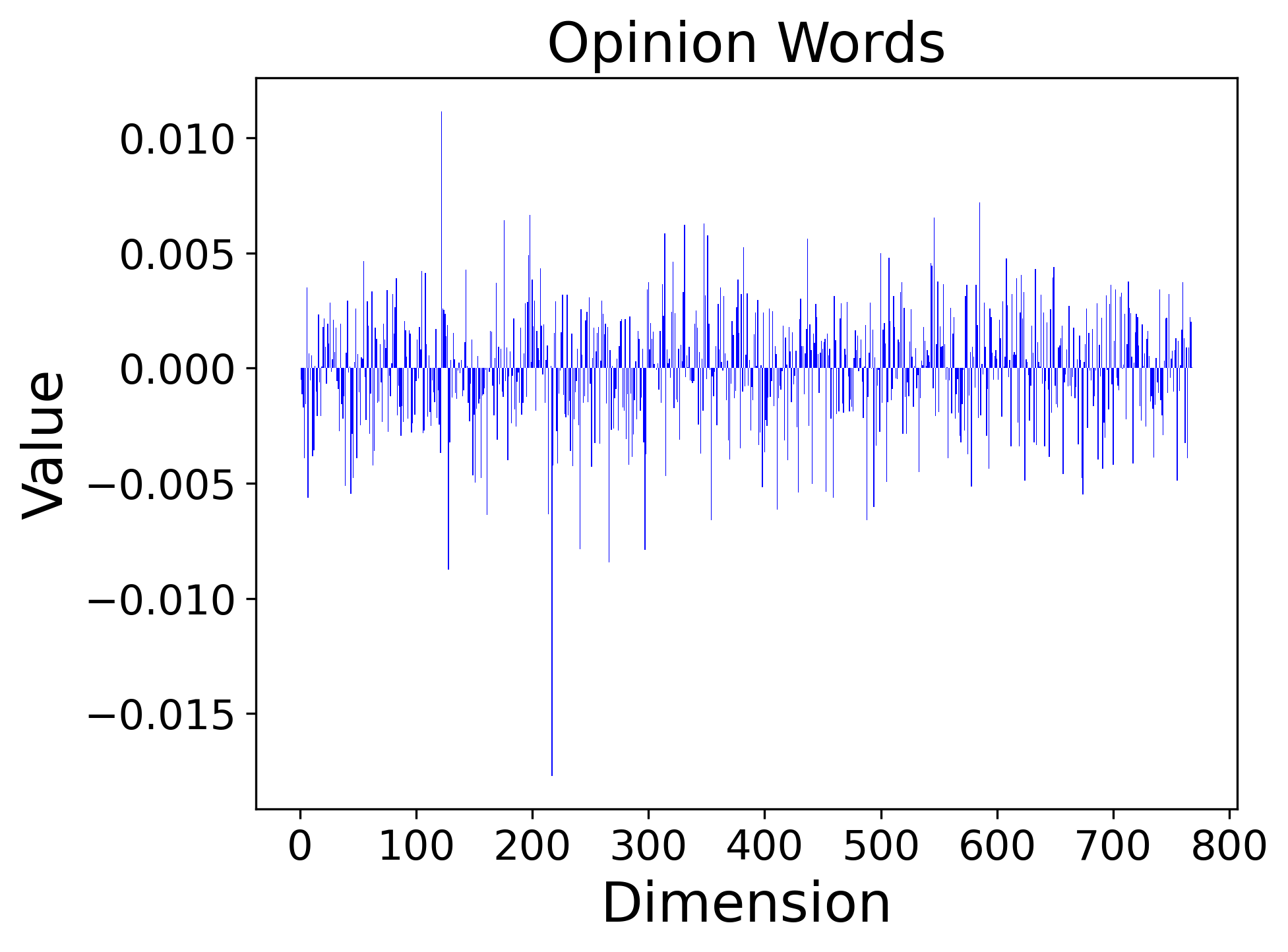}}
\subfigure[Res14] {\includegraphics[scale=0.235]{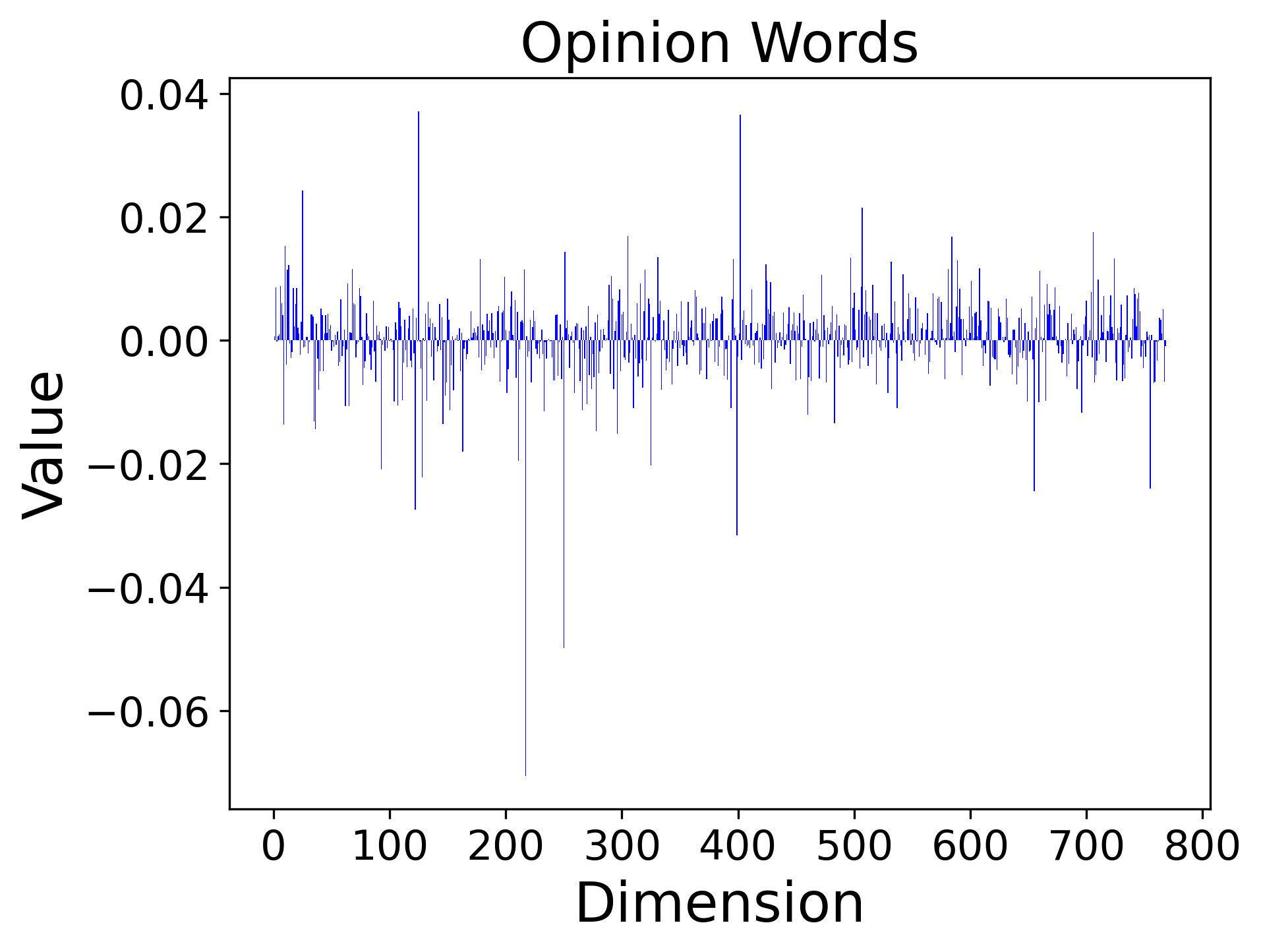}}
\subfigure[Res15] {\includegraphics[scale=0.235]{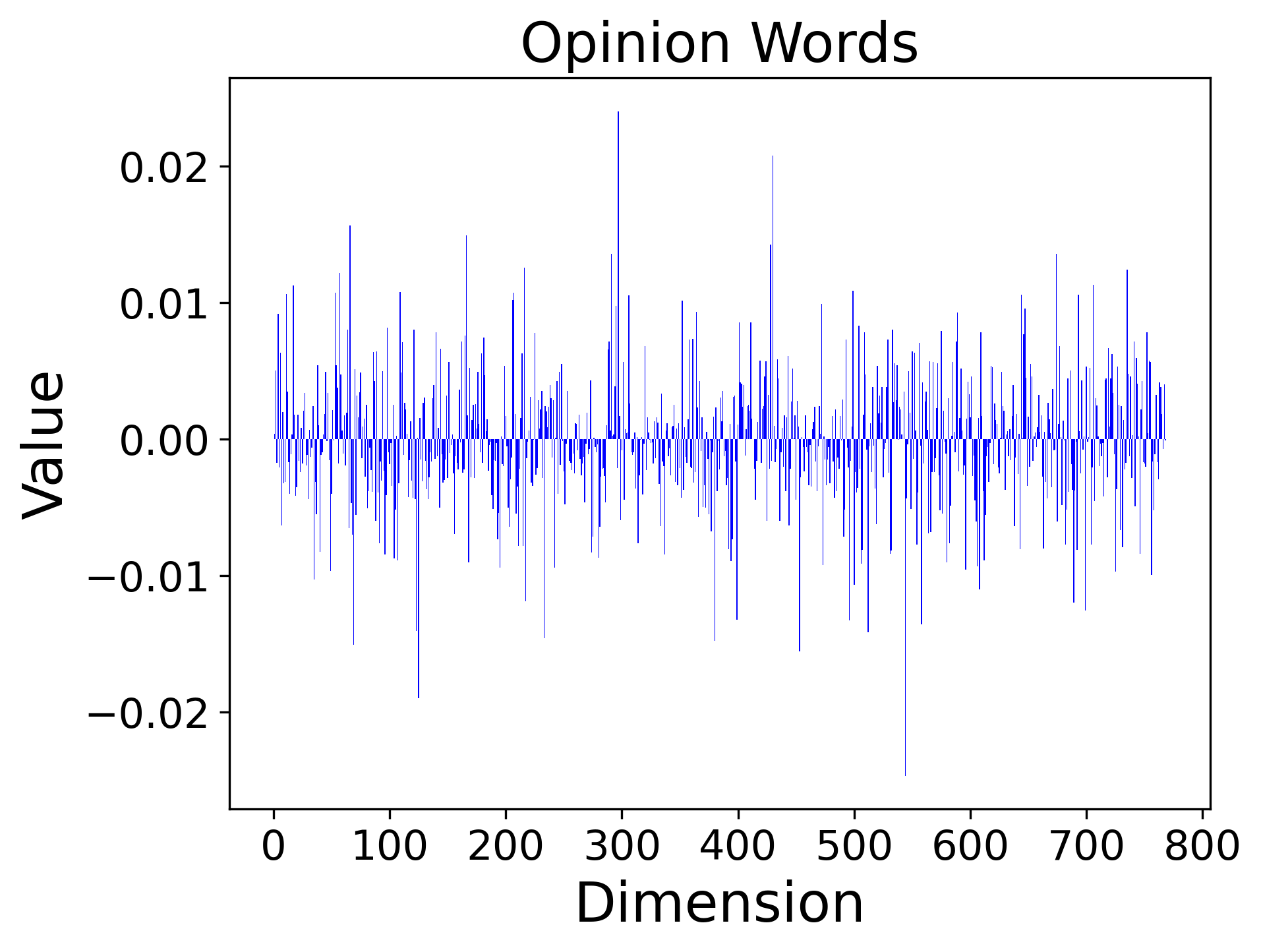}}
\subfigure[Res16] {\includegraphics[scale=0.235]{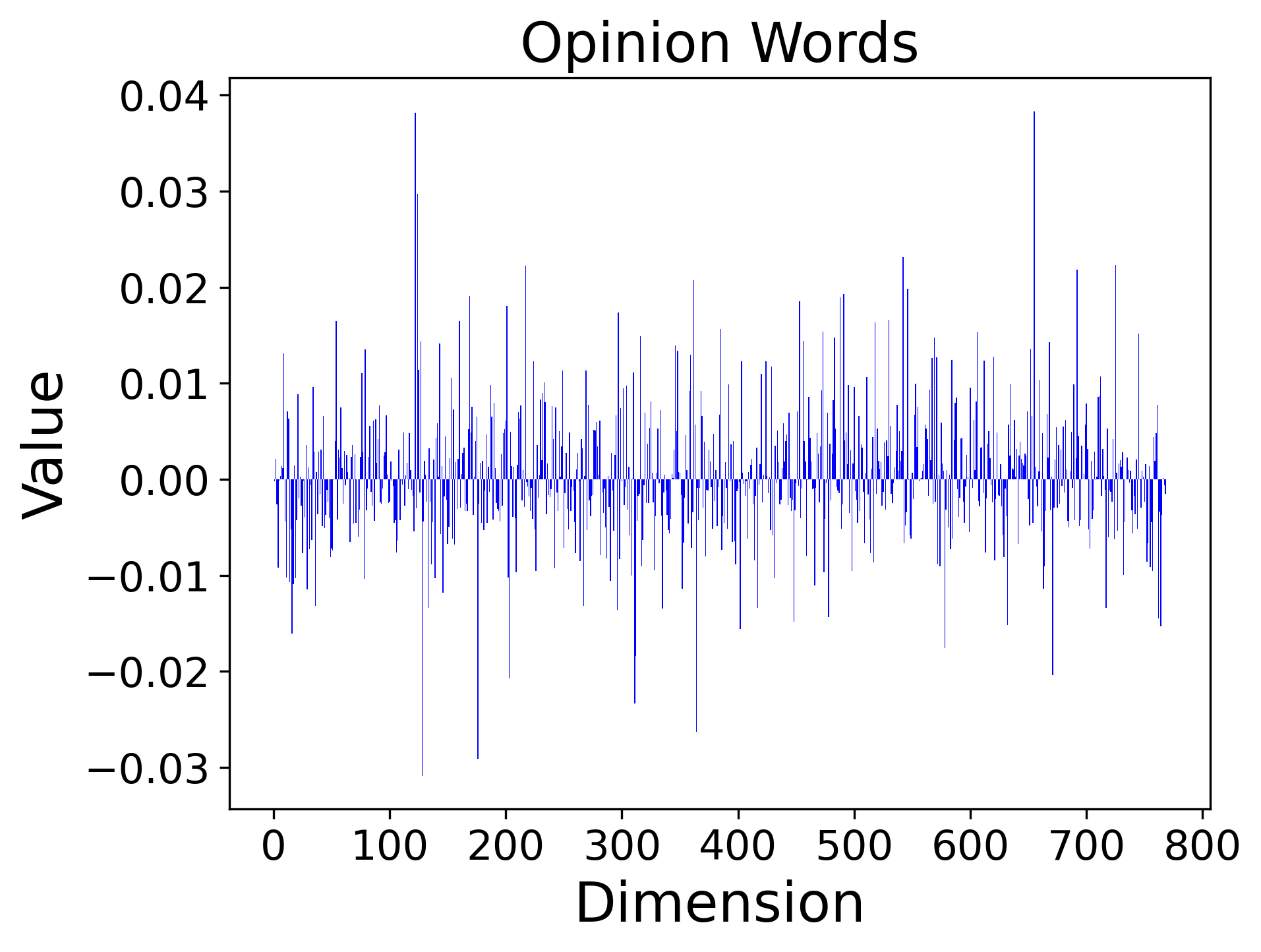}}
\caption{Visualization of the gradients on hidden dimension over four classic ABSA datasets. 
}
  \label{fig:visualization not equally important}
\end{figure*}

In this paper, we aim to answer the question: ``How can important dimensions be dynamically selected?"
We propose an Information Bottleneck-based Gradient explanation framework (\texttt{IBG}) for ABSA to learn the intrinsic dimension. 
To be specific, we propose an Information Bottleneck-based Intrinsic Learning (iBiL) structure to distill word embeddings into an intrinsic dimension that is both concise and replete with pertinent information, ensuring that irrelevant data is judiciously pruned (Figure \ref{fig:intro}). 
Our model is model-agnostic, we integrate it with several state-of-the-art baselines, such as BERT-SPC \cite{kenton2019bert} and DualGCN \cite{li2021dual}.
Our comprehensive evaluations and tests provide substantial evidence of the effectiveness of the \texttt{IBG} framework. As detailed in the following sections, \texttt{IBG} not only enhances the performance metrics but also improves the clarity of interpretations, shedding light on sentiment-aware features.

The key contributions of this paper are listed as follows\footnote{Our code is publicly available at \url{https://github.com/your-username/your-repository}}.
\begin{itemize}[leftmargin=*, align=left]
    \item{We propose the Information Bottleneck-based Gradient (\texttt{IBG}) explanation framework to find the low-dimensional intrinsic space since we discover that not all dimensions of the embedding are equally important in completing the ABSA task through preliminary analyses. 
    }
    \item{We introduce the iBiL structure, forcing the model to learn its intrinsic sentiment embedding by effectively removing irrelevant information while retaining sentiment-related details via information bottleneck. 
    }
    \item{Through extensive experiments, we demonstrate that our framework is capable of enhancing both the performance and the interpretability of the original model significantly.}
\end{itemize}

\begin{figure}[!t]
\centering
\subfigure[Lap14] {\includegraphics[scale=0.20]{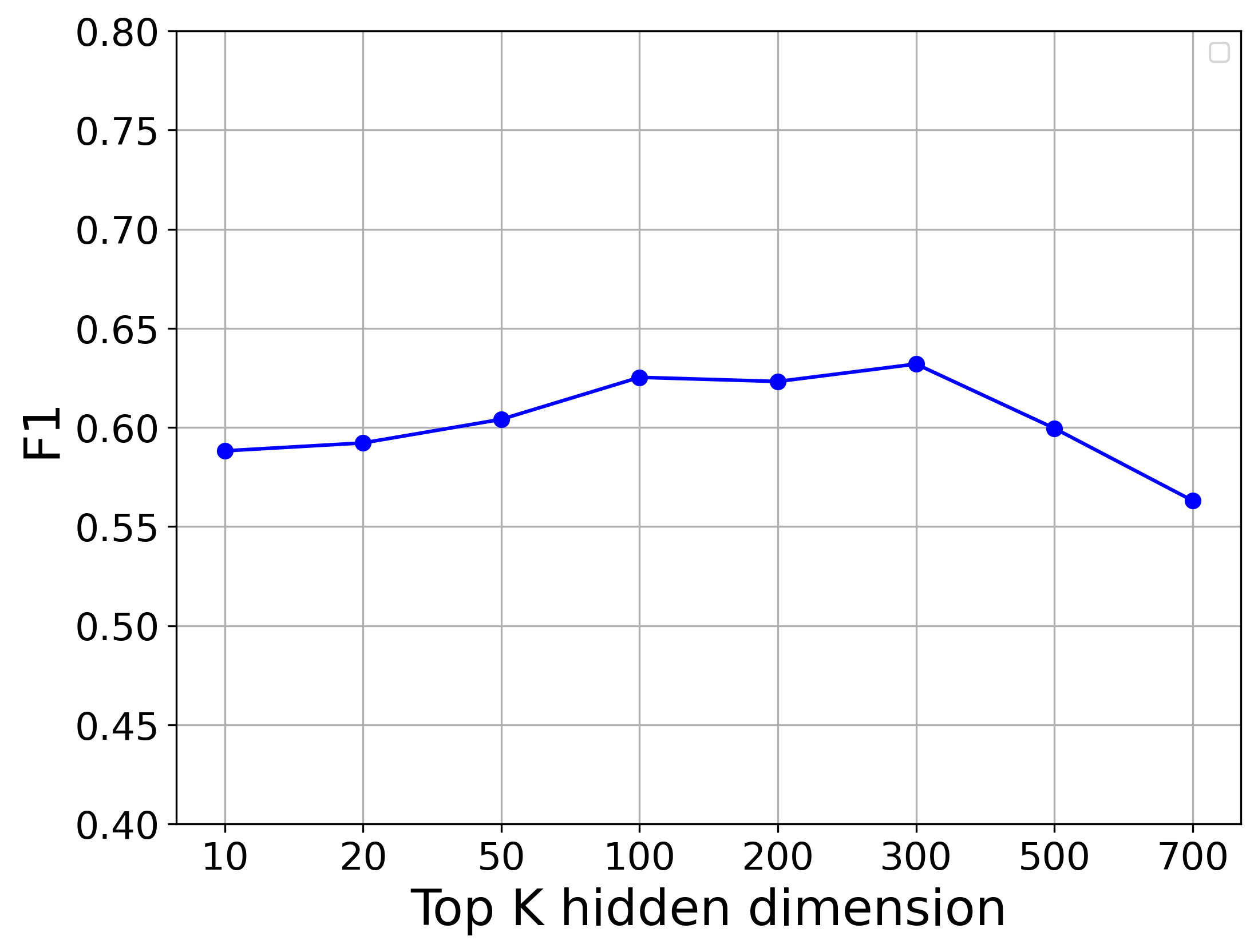}}
\subfigure[Res14] {\includegraphics[scale=0.20]{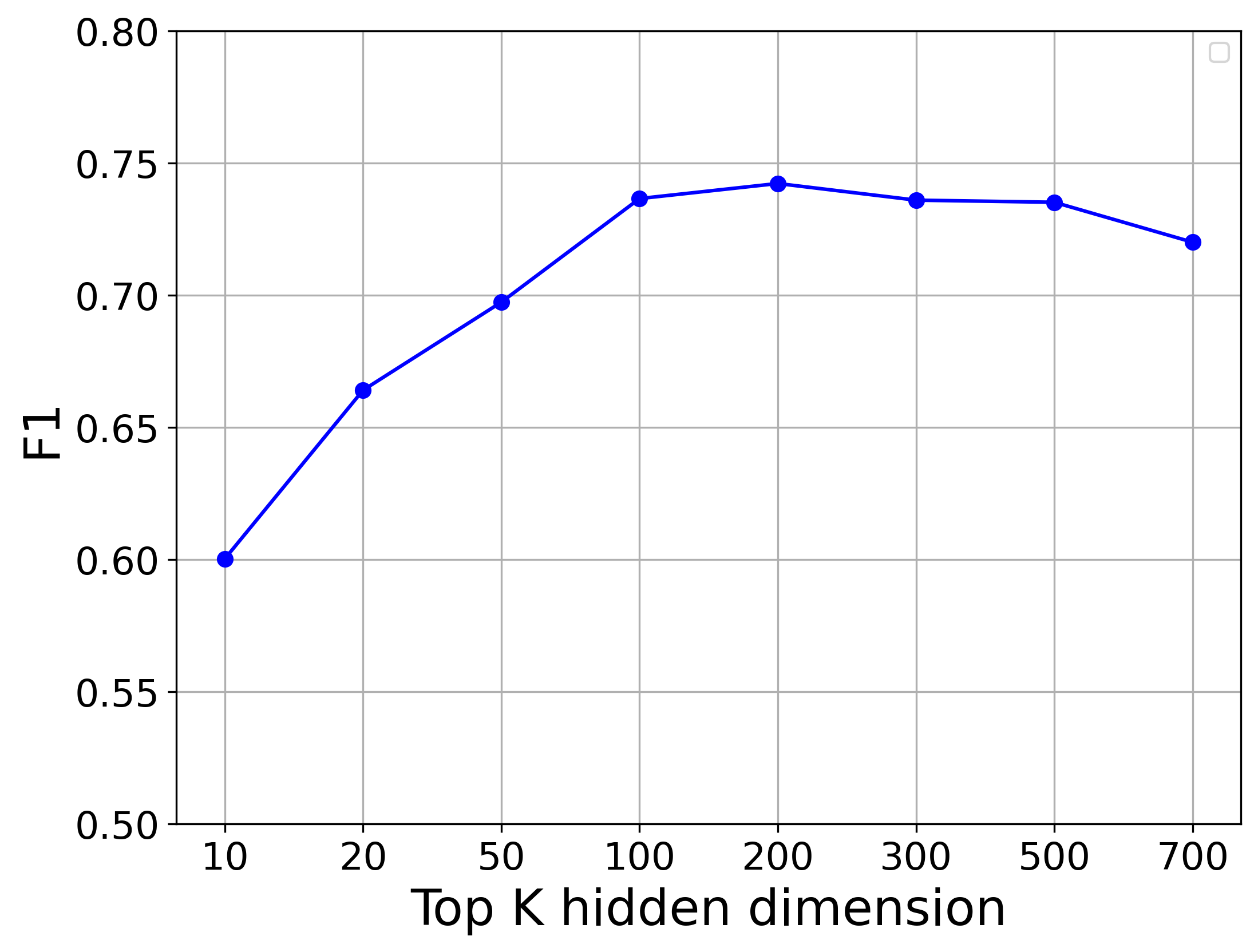}}
\subfigure[Res15] {\includegraphics[scale=0.20]{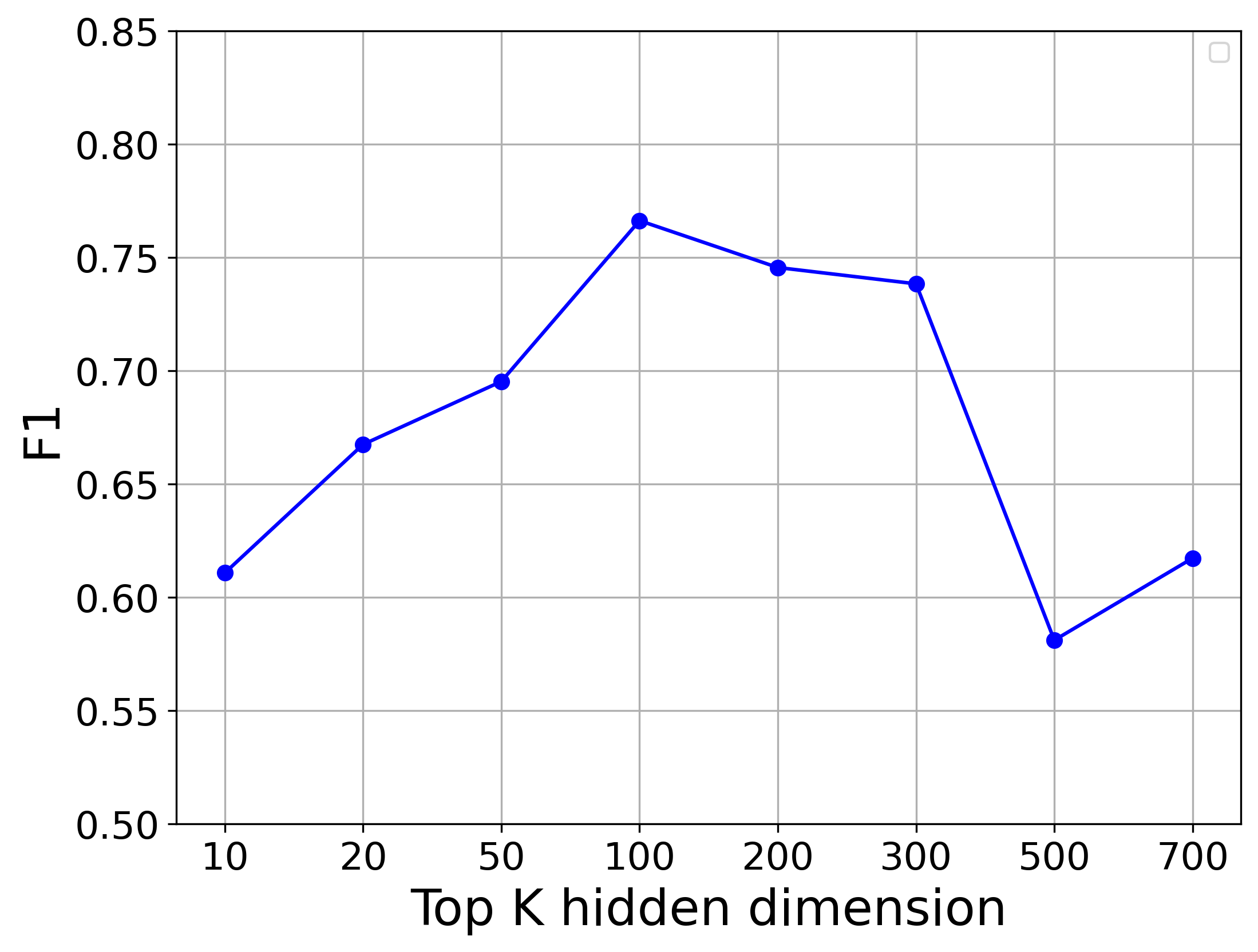}}
\subfigure[Res16] {\includegraphics[scale=0.20]{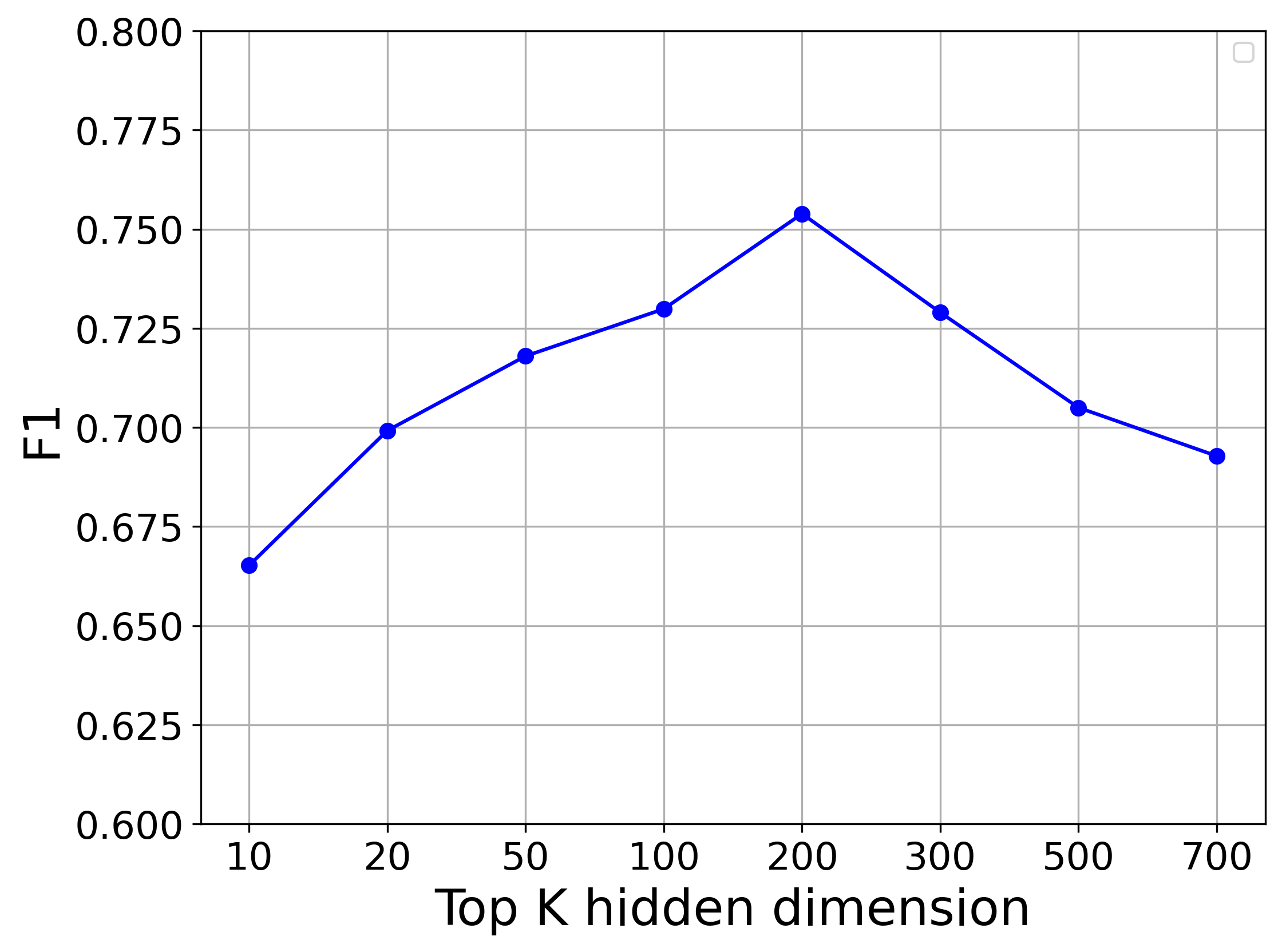}}
\caption{The influence of top k hidden dimension}
  \label{fig:top k}
\end{figure}

\section{Preliminary Analysis}
\label{sect:Preliminary Analysis}
In this section, we mainly conduct preliminary analysis to answer the following three questions.

\textbf{Question 1: Are all the hidden dimensions equally important?}
In Figure \ref{fig:visualization not equally important}, we visualize the gradients of the hidden dimensions. Specifically, we train a sentiment classifier for aspect-based sentiment analysis, which inputs the sentence and aspect into a classifier to predict the sentiment polarity. Then we compute the gradients of the opinion words concerning the given aspects as well as all the words in the sentence. The classifier we used is the Bert-SPC model, which is a very classic and fundamental baseline.

Our observations from these figures can be summarized as follows:
\textbf{First}, we note substantial variations in the values along the dimensions. Several gradient values are notably larger than others, indicating that only a small fraction of dimensions consistently contributes significantly to the model's predictions for each sample. Most dimensions have limited influence.
\textbf{Second}, the values of the opinion words are considerably larger than those of general words. This suggests that gradients can assist in identifying key words.

\textbf{Question 2: How many dimensions are necessary?}
%
The fact that all dimensions do not have equal importance implies that there exist certain dimensions that are really important. Therefore, in this subsection, we will discuss what is the exact number of key dimensions among all dimensions.
Specifically, we employ gradient-based explanation methods to select the top-k important dimensions for predicting sentiment polarity with respect to a given aspect (see Figure \ref{fig:top k}).

Our observations reveal the following:
\textbf{First}, utilizing approximately the top 100-300 dimensions often yields similar results to using all dimensions in most cases.
\textbf{Second}, in certain instances, employing the top-k dimensions can lead to performance improvements compared to using all dimensions.

\begin{figure}[!t]
\centering
\subfigure[Lap14] {\includegraphics[scale=0.14]{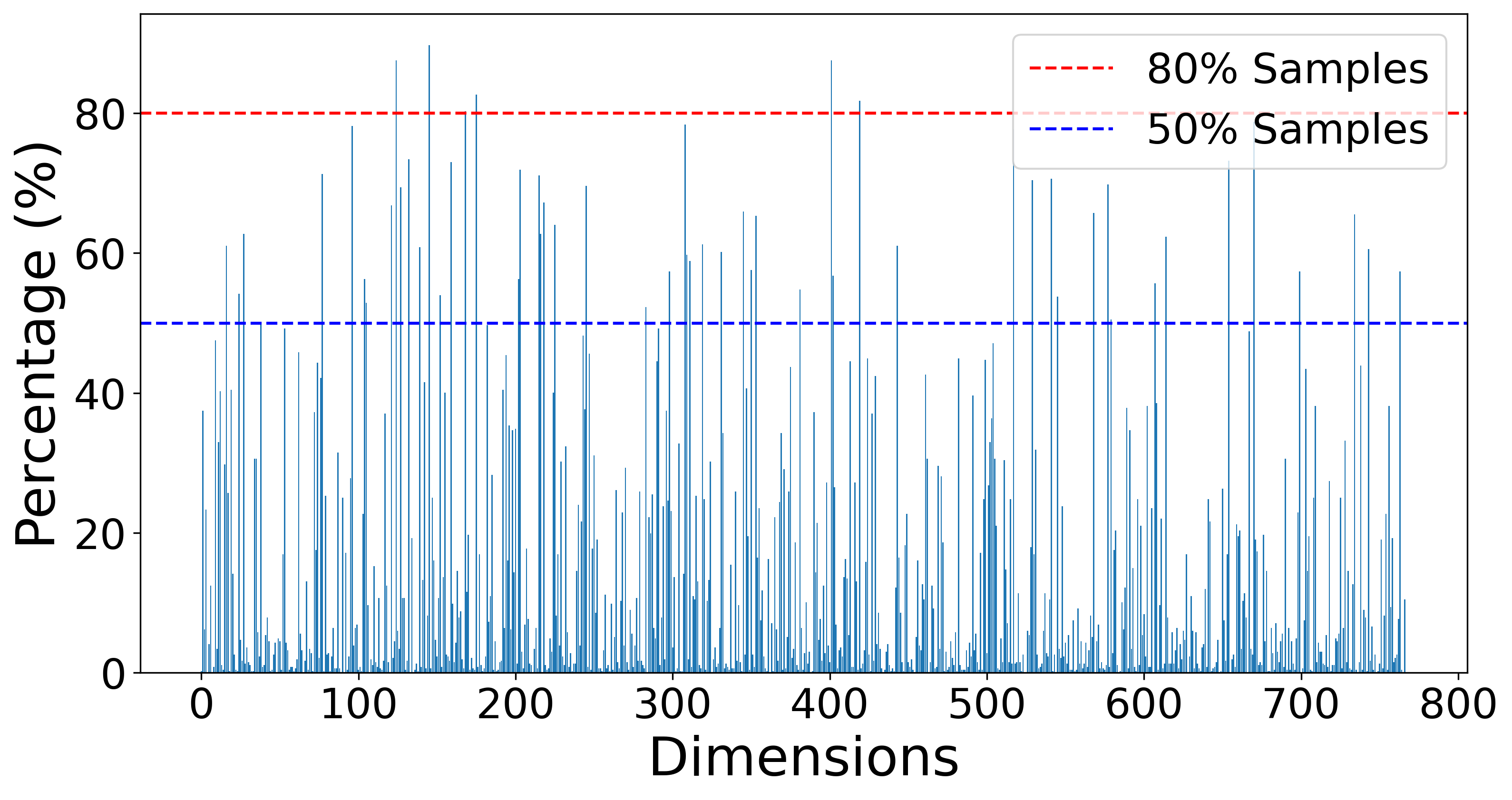}}
\subfigure[Res14] {\includegraphics[scale=0.14]{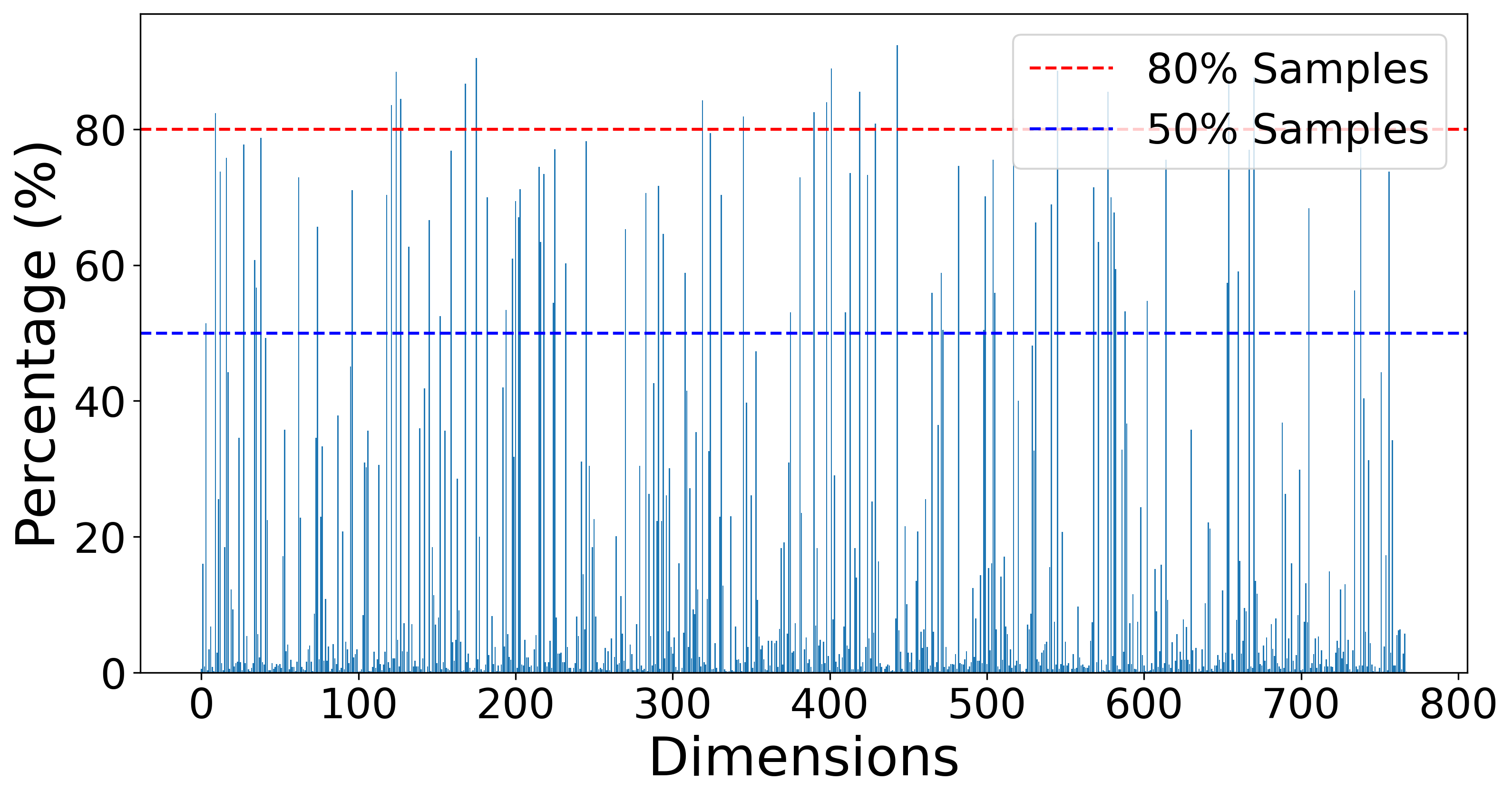}}
\subfigure[Res15] {\includegraphics[scale=0.14]{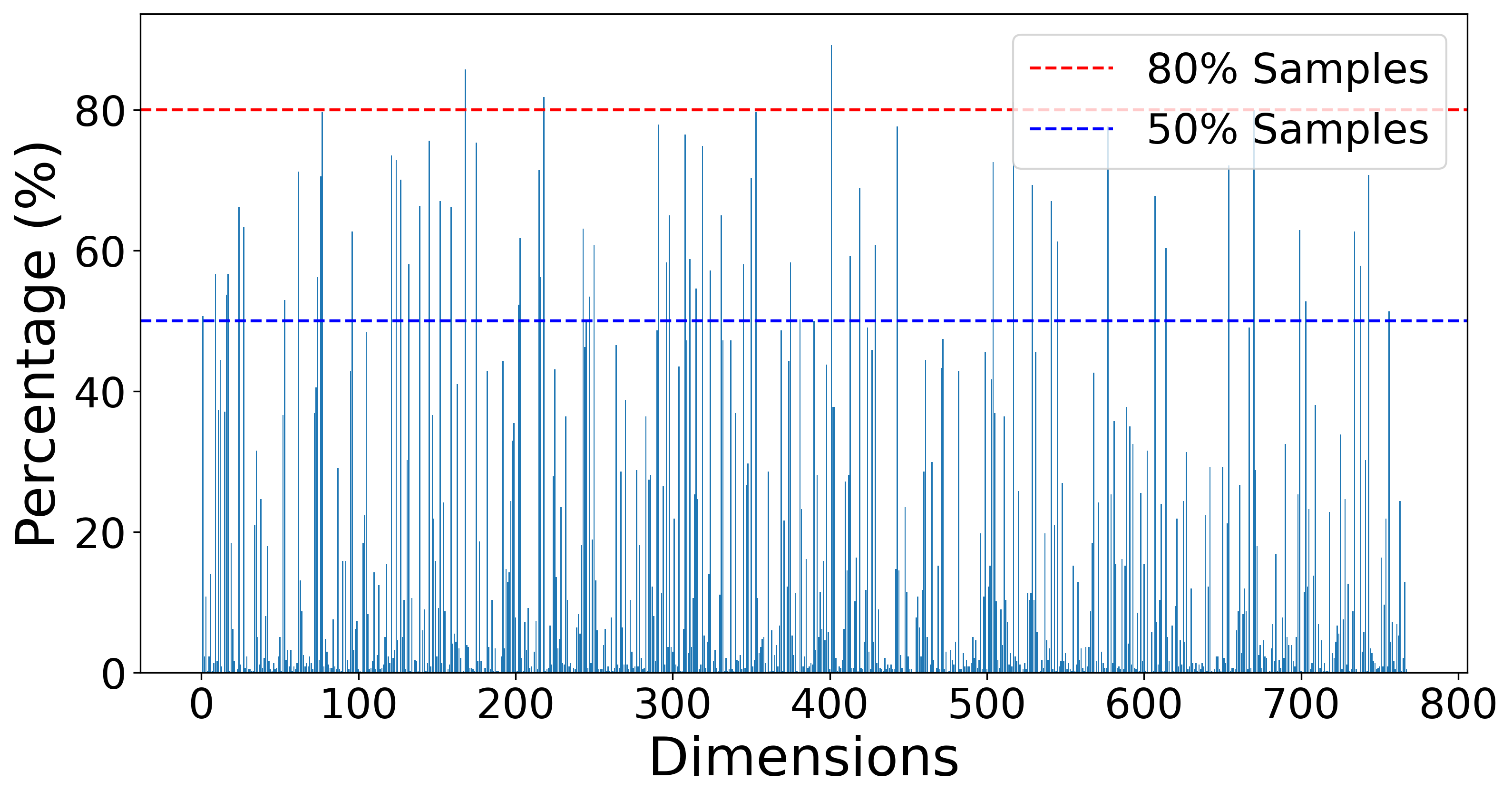}}
\subfigure[Res16] {\includegraphics[scale=0.14]{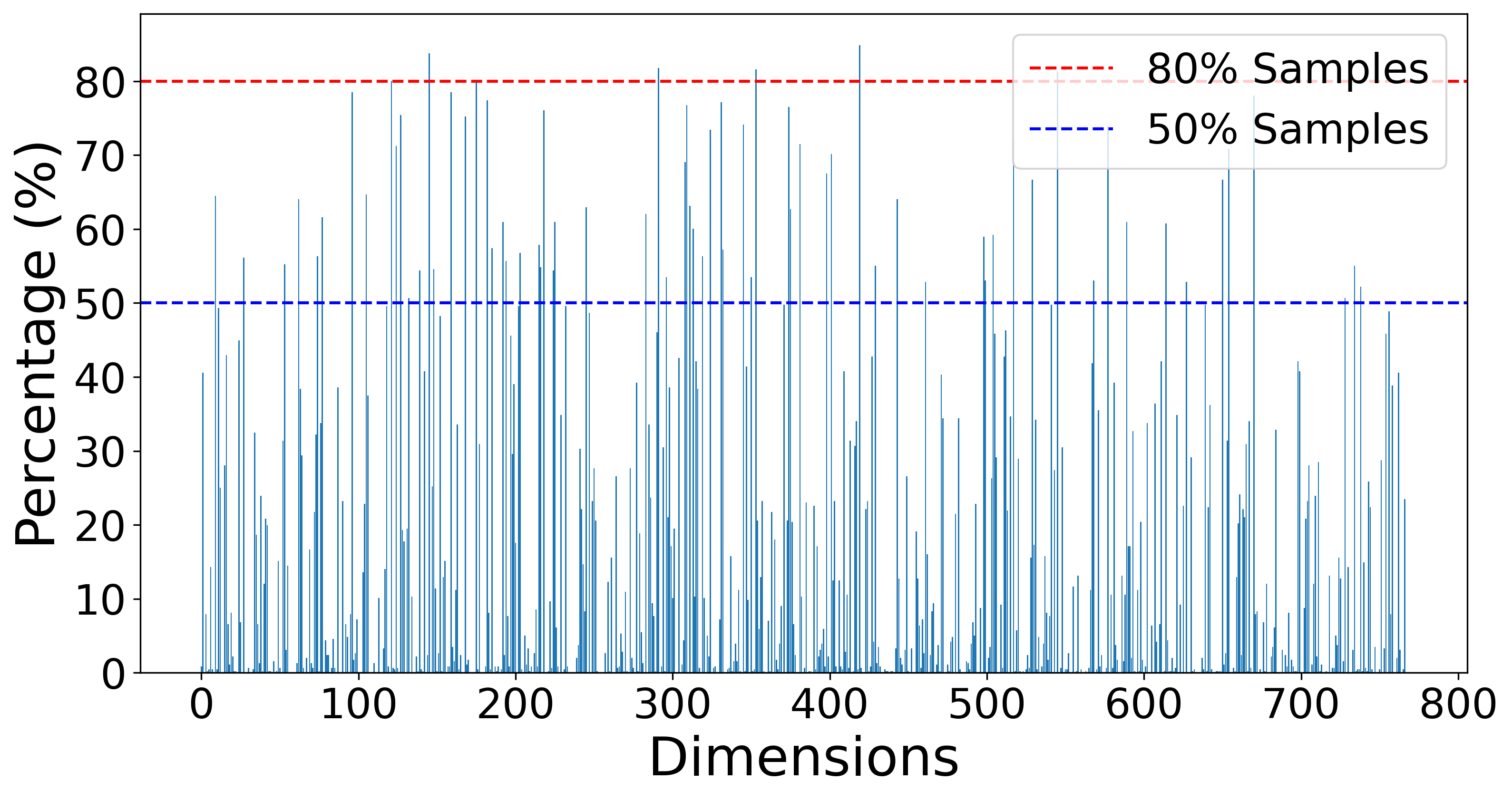}}
\caption{Percentage of Samples Where Each Dimension is in the Top 100 Importance}
  \label{fig:good_dimensions_frequency}
\end{figure}

\textbf{Question 3: Which dimensions are important?}
%
Therefore, aiming to ascertain whether certain dimensions consistently play a crucial role in ABSA, we conduct a statistical analysis focusing on the top-100 important dimension indices in each sample's prediction, as detailed in Figure \ref{fig:good_dimensions_frequency}.

Our findings indicate the following:
\textbf{First}, it is noteworthy that across more than 50\% of samples within each dataset, approximately 80 dimensions consistently maintain their positions within the top-100 important dimension indices.
\textbf{Second}, several dimension indices consistently demonstrate significance in nearly every sample within a dataset. 
For instance, the top-3 significant dimensions are (145, 124, 401), (443, 175, 401), (401, 168, 218) and (419, 145, 219) in the Lap14, Res14, Res15 and Res16 datasets, respectively.
However, it is essential to recognize that the specific important dimensions vary among different datasets. 


\begin{figure*}[!ht]
\begin{center}
\includegraphics[scale=0.6]{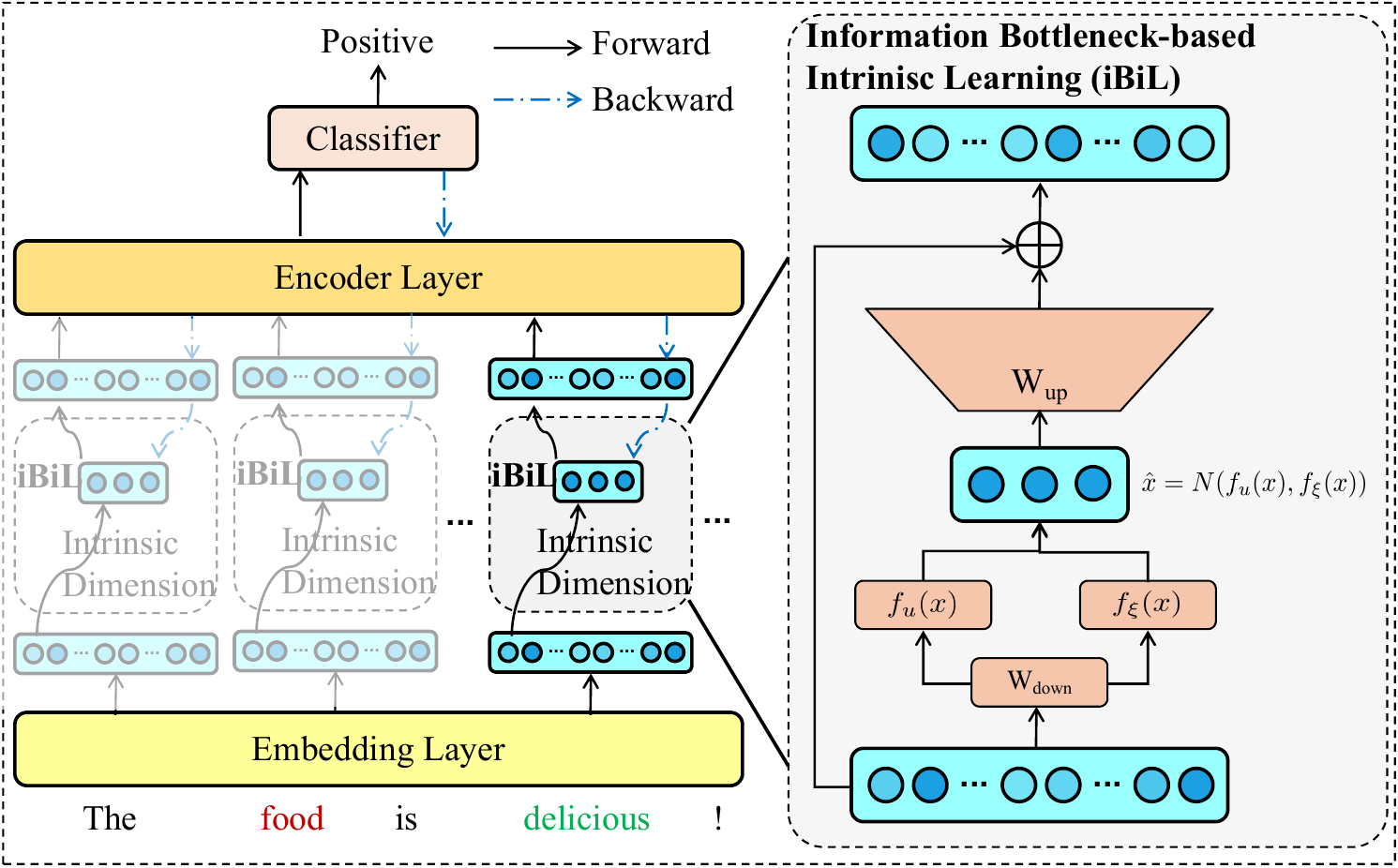} 
\caption{The framework of our \texttt{IBG} approach.}
\label{fig:framework}
\end{center}
\end{figure*}

\section{Our Approach}
In this paper, we propose the \texttt{IBG} explanation framework for ABSA by learning the low-dimensional intrinsic features via information bottleneck (Figure \ref{fig:framework}). 
Our framework explains the sentiment classifier by extracting the aspect-aware opinion words via the gradient method (Section \ref{sect:overview}). 
It calculates the important weights according to the gradients on the embedding level, which contains redundancy information.
Thus, we introduce our self-designed Information Bottleneck-based Intrinsic Learning (iBiL) structure between the embedding layer and encoder layer of the traditional language model (Section \ref{sect:ibil}). 
This incorporation utilizes the information bottleneck principle to compress the embedding layer to get the intrinsic representations, thereby eliminating redundant information in the original embedding and retaining essential information.

Let $s$ be a sentence with words $\{w_1, w_2, ..., w_{|s|}\}$, $a$ be an aspect in the sentence $s$, and $y \in Y$ be the sentiment label of $a$.
Given a corpus 
$\mathcal{D}=\{(s_i, a_i, y_i)\}_{i=1}^{|\mathcal{D}|}$,
our task is to predict the sentiment polarity $y \in \{P, N, O\}$ of the sentence towards the given aspect $a$. $P$, $N$, $O$ represent positive, negative and neutral, respectively. 
The word embeddings of the sentence $s$ are $x=\{x_1, ..., x_i, ..., x_{|s|}\}$, where $x_i$ is the word embedding of $w_i$. Moreover, we aim to explain the model by extracting the \emph{aspect-aware opinion words} $o$ that express the sentiment w.r.t the aspect $a$.


\subsection{Overview of \texttt{IBG}}
\label{sect:overview}
In this section, we introduce the overall structure of our \texttt{IBG}. 
It is a gradient-based explanation method based on an aspect-based sentiment classifier, which computes the importance scores for each token based on the embedding to find the aspect-specific opinion words (e.g., delicious)
To improve the interpretability, it incorporates our designed iBiL structure between the embedding and encoder layers. 
This structure is employed to acquire intrinsic representations at the embedding level, eliminating redundant information while retaining emotion-related essential information.

The prevailing paradigm of gradient-based model interpretation methods in NLP consists of two main steps. First, a pre-trained language model, including the embedding layer, is fine-tuned for specific downstream tasks. Subsequently, importance scores for each token are computed, primarily through the embedding layer.

Particularly, we train a sentiment classifier using our iBiL for ABSA, which aims to predict the sentiment of the sentence concerning the given aspect. 
Let $\mathcal{F}$ be a sentiment classifier with embedding and encoder layers that predicts the sentiment distribution $P(y|s, a)$ based on the sentence $s$ and aspect $a$.
\begin{equation}
    P(y|s,a) = \mathcal{F}(s, a)
\end{equation}
It is worth noting that, given our framework's model-agnostic nature, the sentiment classifier $\mathcal{F}$ can be any existing ABSA model. In this paper, we mainly conduct the experiments on BERT-SPC \cite{kenton2019bert} and DualGCN-BERT \cite{li2021dual}.

After the words $w$ in the sentence $s$ passing through the embedding layer, we obtain embeddings $x=\{x_1, ..., x_i, ..., x_{|s|}\}$, where $x_i \in \mathcal{R}^{\mathrm{High}}$ is the high dimension embedding of word $w_i$. 
Afterward, we employ iBiL structures to learn distinct intrinsic dimensional representations via the information bottleneck structure, which will be introduced in Section \ref{sect:ibil}. 
Through iBiL, we also obtain the ``intrinsic representation" $\hat{x}=\{\hat{x}_1, ..., \hat{x}_i, ..., \hat{x}_{|s|}\}$,  where $\hat{x}_i \in \mathcal{R}^{\mathrm{Low}}$ is the low dimension representation of word $w_i$. 
Here, $\mathrm{High}$ is the dimensionality of the hidden layer in the original embedding, while $\mathrm{Low}$ represents the size of the intrinsic dimensionality, where $\mathrm{High}$ (e.g., 768) is much larger than $\mathrm{Low}$ (e.g., 10, 20). 
To input the final word representation $x^{\prime}$ into the encoder layer, we upsample the intrinsic representation into a hidden dimension.
\begin{equation}
    x^{\prime} = x + W_{up}\hat{x}
\end{equation}
where $W_{up}$ represents a learnable upsampling matrix. Throughout the entire training process, $x$ remains frozen and unchanged.


Subsequently, a novel attribution method is employed to determine the importance scores for each token based on both the original embeddings and intrinsic representations, ultimately leading to improvements in models' performance and interpretability.
Similar to prior work \cite{DBLP:journals/corr/SimonyanVZ13}, we compute importance scores by dotting the gradients on the model's prediction with the corresponding vectors:
\begin{equation}
\begin{split}
{\gamma}(w_i) &= \left|x_i \times \frac{\partial \mathcal{F}(s, a)}{\partial x_i}\right| \\
\hat{\gamma}(w_i) &= \left|\hat{x}_i \times \frac{\partial \mathcal{F}(s, a)}{\partial \hat{x}_i}\right|
\end{split}
\end{equation}
where $\gamma$ and $\hat{\gamma}$ are the importance weights for each token in the embedding layer and the intrinsic representation layer.

Finally, for each token's $\gamma$ and $\hat{\gamma}$, we introduce a hyperparameter $\alpha$ to balance the weights between the two scores. For each word $w_i$, the final importance score is calculated as:
\begin{equation}
\label{alpha equation}
\mathrm{FScore}(w_i) = (1-\alpha)\gamma(w_i) + \alpha\hat{\gamma}(w_i)
\end{equation}

Finally, we use $\mathrm{FScore}(w_i)$ to select the opinion words of aspect.

\subsection{Information Bottleneck-based Intrinsic Learning}
\label{sect:ibil}
In our proposed \texttt{IBG} framework, one of the key innovations lies in the design of Information Bottleneck-based Intrinsic Learning (iBiL). This involves compressing the pre-trained embedding $x$ into a lower-dimensional space $\hat{x}$ to learn the intrinsic sentiment representation in the embedding layer. 
The primary goal is to retain essential information in the embedding while removing redundant noise.

According to the original IB theory \cite{alemi2016deep}, the main objective is to learn a compressed representation $Z$ by maximizing the mutual information between $Z$ and output $Y$ and minimizing the mutual information between $Z$ and input $X$. 
In our work, to learn the intrinsic representation $\hat{x}$, we aim to preserve the information of $\hat{x}$ with respect to the sentiment polarity $y$ while minimizing the mutual information between $\hat{x}$ and the original word vectors $x$. Therefore, our goal is to minimize the following loss:
\begin{equation}
    \mathcal{L} = \beta I(x;\hat{x}) - I(y;\hat{x}) 
\label{classic ib formula}
\end{equation}
where $I(.,.)$ represents the mutual information.

However, in practical operations, directly calculating the mutual information $x$ and $\hat{x}$ is not feasible. We employ the Variational Inference method \cite{DBLP:conf/emnlp/LiE19} to estimate the final loss function. Specifically, the upper bound of $I(x; \hat{x})$ can be obtained through the following method,
\begin{equation}
\begin{split}
    I(x;\hat{x}) \leq \int \, d\hat{x}dxp(x)p(x,\hat{x})\log \frac{p(\hat{x} \mid x)}{q(\hat{x})}
\end{split}
\label{upper bound}
\end{equation}

Since
\begin{equation}
\begin{aligned}
    & KL[p(\hat{x}),q(\hat{x})] \geq 0 \\
    \implies & \int \, d\hat{x}p(\hat{x})\log p(\hat{x}) \geq \int \, d\hat{x}p(\hat{x})\log q(\hat{x})
\end{aligned}
\end{equation}

In actual computation process, we first sample $\hat{x}$ from the original word vectors using the reparameterization method \cite{DBLP:journals/corr/KingmaW13}:
\begin{equation}
    p(\hat{x} \mid x) = f_\mu(x) + f_\xi(x) \cdot z
\end{equation}
where $z \sim \mathcal{N}(0, I)$. Here, $f_\mu$ and $f_\xi$ correspond to two linear layers, which are trainable. And $q(\hat{x})$ in Equation \ref{upper bound} is assumed to follow the standard normal distribution.
Therefore, $I(x; \hat{x})$ in Equation \ref{classic ib formula} can be directly replaced with the KL loss in Equation \ref{upper bound} and be easily calculated.

The lower bound of $I(y; \hat{x})$ can also be obtained as follows,
\begin{equation}
\small
\begin{split}
    I(y;\hat{x}) \geq \int \, dyd\hat{x}p(y,\hat{x})\log q(y \mid \hat{x}) - \int \, dyp(y)\log p(y)
\end{split}
\label{lower bound}
\end{equation}

Since
\begin{equation}
\begin{aligned}
    KL[p(y \mid \hat{x}),q(y \mid \hat{x})] \geq 0 &\implies \\
    \int \, dy p(y \mid \hat{x})\log p(y \mid \hat{x}) \geq &\int \, dy p(y \mid \hat{x})\log q(y \mid \hat{x})
\end{aligned}
\end{equation}

The $\int \, dyp(y)\log p(y)$ in Equation \ref{lower bound} is a constant, and thus, we do not need to pay attention to it during the optimization.
If we consider $q(y \mid \hat{x})$ as the subsequent dimensionality-expanding matrix and encoder structure of the model, then the process of minimizing $I(y; \hat{x})$ can be directly approximated as optimizing the original sentiment classification loss function. In other words, $I(y; \hat{x})$ can be straightforwardly replaced by a cross-entropy loss $\mathcal{L}_{CE}$.
The final loss function can be written as follows.
\begin{equation}
    \mathcal{L} = \mathcal{L}_{CE} + \beta 
 KL[p(\hat{x} \mid x), q(\hat{x})]
\end{equation}

\section{Experiment Setups}
\subsection{Datasets, Metrics and Settings}
\paragraph{Datasets.} To evaluate the performance and the interpretability of our \texttt{IBG} framework, we conduct a comprehensive series of experiments on four common datasets: Res14, Lap14, Res15 and Res16 \cite{fan-etal-2019-target}, which labeled the opinion words for each aspect. 

\paragraph{Metrics.} In order to assess the performance of our framework, we employ the most widely used metrics in the ABSA task: Accuracy and Macro F1-score.
Moreover, to verify the effectiveness of our framework in improving models' interpretability, following \cite{chen2020learning}, we adopt the area over the perturbation curve (AOPC) \cite{nguyen2018comparing,samek2016evaluating} and Post-hoc Accuracy (Ph-Acc) \cite{chen2018learning}. 
AOPC computes the average decrease in accuracy when the model makes predictions after removing the top-k important words for explanation. Ph-acc, on the other hand, retains only the top-k words and masks the remaining words to assess whether the model can still make accurate predictions.

\paragraph{Settings.} We use bert-base-uncased version \cite{kenton2019bert} and Adam optimizer for the original BERT and the additional components we introduce with learning rates 1e-5 and 1e-4, respectively. In the selection of intrinsic dimensionality sizes, we consider dimensions of 5, 10, 20, 50, 100 and 300. The hyperparameter $\alpha$ was set to 0.5.

\begin{table*}[t!]
\centering
\scriptsize
\setlength{\tabcolsep}{0.5mm}{\begin{tabular}{l|cccc|cccc|cccc|cccc}
\hlineB{4}
\multirow{2}{*}{Model}    & \multicolumn{4}{c|}{Lap14} & \multicolumn{4}{c|}{Res14} & \multicolumn{4}{c|}{Res15} & \multicolumn{4}{c}{Res16} \\
& Acc & F1 & AOPC & Ph-Acc & Acc & F1 & AOPC & Ph-Acc & Acc & F1 & AOPC & Ph-Acc & Acc & F1 & AOPC & Ph-Acc \\
\hline
AEN-BERT & 81.80 & 56.07 & -  & -  & 88.59 & 72.69 & -  & - & 86.44 & 63.73 & -  & - &88.60 & 65.06 & -  & - \\
LCF-BERT & 81.83 & 58.23 & - & - & 90.00 & 67.91 & - & - & 85.94 & 67.53 & - & - & 89.91 & 69.98 & - & - \\
ASCM4ABSA & 81.93 & 57.34 & - & - & 89.96 & 70.85 & - & - & 86.81 & 66.32 & - & - & 88.98 & 68.04 & - & - \\
ChatGPT & 82.78 & 44.60 & - & - & 90.62 & 51.72 & - & - & 90.54 & 71.86 & - & - & 93.63 & 76.70 & - & - \\
BERT-SPC$_\texttt{IEGA}$ & 82.28 & 62.93 & 15.04 & 42.18 & 90.62 & 72.75 & 11.13 & 69.18 & 85.40 & 59.39 & 08.26 & 70.78 & 88.56 & 62.60 & 10.48 & 75.49 \\
RGAT-BERT$_\texttt{IEGA}$ & 82.58 & 65.10 & 13.58 & 66.52 & 91.64 & 77.50 & 15.67 & 63.29 & 87.09 & 69.36 & 16.07 & 72.98 & 90.78 & 67.34 & 12.71 & 80.04 \\
BERT-SPC$_\texttt{Grad}$ & 81.80 & 56.31 & 11.13 & 67.02 & 90.47 & 72.01 & 09.88 & 62.71 & 88.02 & 61.73 & 10.60 & 76.73 & 90.45 & 70.32 & 10.53 & 79.17 \\
BERT-SPC$_\texttt{InteGrad}$ & - & - & 09.85 & 74.09 & - & - & 10.47 & 54.24 & - & - & 16.59 & 80.88 & - & - & 10.31 & 81.40 \\
BERT-SPC$_\texttt{SmoothGrad}$ & - & - & 09.21 & 69.81 & - & - & 12.59 & 69.65 & - & - & 12.67 & 78.11 & - & - & 09.43 & 81.58 \\
DualGCN-BERT$_\texttt{Grad}$ & 84.27 & 67.07 & 18.84 & 64.60 & 91.41 & 76.62 & 11.65 & 77.65 & 88.71 & 69.82 & 13.36 & 73.04 & 91.67 & 76.97 & 13.60 & 73.46 \\
DualGCN-BERT$_\texttt{InteGrad}$ & - & - & 14.35 & 59.46 & - & - & 13.06 & 74.35 & - & - & 14.29 & 74.19 & - & - & 14.91 & 71.71 \\
DualGCN-BERT$_\texttt{SmoothGrad}$ & - & - & 22.70 & 68.03 & - & - & 15.53 & 80.24 & - & - & 14.29 & 74.19 & - & - & 16.23 & 75.00 \\
\hline
BERT-SPC$_\texttt{IBG}$ & 83.30 & 65.34 & 17.77 & \textbf{75.16} & 91.29 & 76.33 & \textbf{28.35} & \textbf{84.24} & 89.40 & \textbf{78.32} & 17.28 & \textbf{81.34} & 92.32 & 79.57 & 20.39 & \textbf{83.33} \\
DualGCN-BERT$_\texttt{IBG}$ & \textbf{85.22} & \textbf{72.99} & \textbf{26.89} & 66.17 & \textbf{92.58} & \textbf{80.22} & 27.18 & 78.71 & \textbf{90.63} & 77.68 & \textbf{25.58} & 75.81 & \textbf{93.64} & \textbf{84.61} & \textbf{20.83} & 82.24  \\
\hlineB{4}
\end{tabular}}
\caption{The main results of performance and interpretability.}
\label{table:main table}
\end{table*}

\subsection{Baselines}
We compare our framework with the SOTA baselines to investigate its performance and interpretability.
To validate the performance, we select the following baselines:
\begin{itemize}[leftmargin=*, align=left]
    \item BERT-SPC \cite{kenton2019bert} simply concatenates the raw sentences with the corresponding aspect terms, subsequently feeding these inputs directly into a pre-trained BERT model for ABSA.
    \item AEN-BERT \cite{song2019attentional} proposes an Attentional Encoder Network (AEN) and enhances BERT with attention-based encoders to capture context-specific information related to aspects.
    \item LCF-BERT \cite{zeng2019lcf} designs a Local Context Focus (LCF) mechanism which uses multi-head self-attention to force the model to pay attention to the local context words.
    \item RGAT-BERT \cite{wang2020relational} is the first work that uses the Graph Convolutional Network (GCN) in ABSA to utilize the syntactical dependency structures in the sentences.
    \item DualGCN-BERT \cite{li2021dual} is a Dual GCN, employing two GCNs, which can simultaneously capture both syntax and semantic information.
    \item ASCM4ABSA \cite{ma2022aspect} proposes three aspect-specific input methods and exploits these transformations to promote the language models to pay more attention to the aspect-specific context in ABSA. 
    \item ChatGPT \cite{wang2023chatgpt} designs a prompt specifically for ABSA, using the GPT-3.5-turbo model to generate the results and analyze its performance.
    \item IEGA \cite{cheng2023tell} is a model agnostic Interpretation-Enhanced Gradient-based framework for ABSA, which guides the model's attention towards important words like opinion words.
\end{itemize}

In addition, to verify the interpretability, we select some classic gradient-based explanation strategies for comparison. 
\begin{itemize}[leftmargin=*, align=left]
    \item Simple Gradient \cite{DBLP:journals/corr/SimonyanVZ13} calculates the gradient of the model's output w.r.t. the input by taking the dot product of these gradients with the corresponding feature values.
    \item Smooth Gradient \cite{DBLP:journals/corr/SmilkovTKVW17} is an enhancement of the Simple Gradient that reduces noise and provides more stable explanations by computing the gradient at multiple points along the path to the actual input.
    \item Integrated Gradients \cite{sundararajan2017axiomatic} considers a baseline input and calculates the gradient at multiple points along the straight-line path, connecting the baseline to the actual input.
\end{itemize}

\begin{table*}[t!]
\centering
\scriptsize
\setlength{\tabcolsep}{0.8mm}{\begin{tabular}{l|cccc|cccc|cccc|cccc}
\hlineB{4}
\multirow{2}{*}{Model}    & \multicolumn{4}{c|}{Lap14} & \multicolumn{4}{c|}{Res14} & \multicolumn{4}{c|}{Res15} & \multicolumn{4}{c}{Res16} \\
& Acc & F1 & AOPC & Ph-Acc & Acc & F1 & AOPC & Ph-Acc & Acc & F1 & AOPC & Ph-Acc & Acc & F1 & AOPC & Ph-Acc \\
\hline
BERT-SPC$_\texttt{IBG}$ & 83.30 & 65.34 & 17.77 & 75.16 & 91.29 & 76.33 & 28.35 & 84.24 & 89.40 & 78.32 & 17.28 & 81.34 & 92.32 & 79.57 & 20.39 & 83.33 \\ \hline
w/o IB & 82.87 & 64.66 & 18.63 & 71.09 & 90.94 & 74.62 & 28.94 & 72.71 & 89.40 & 76.14 & 17.28 & 78.57 & 91.45 & 78.54 & 17.76 & 82.46 \\
w/o iBiL & 81.80 & 56.31 & 11.13 & 67.02 & 90.47 & 72.01 & 09.88 & 62.71 & 88.02 & 61.73 & 10.60 & 76.73 & 90.45 & 70.32 & 10.53 & 79.17 \\
\hlineB{4}
\end{tabular}}
\caption{The results of ablation studies.}
\label{table:ablation study}
\end{table*}
\section{Results and Analyses}
\subsection{Main Results}

To demonstrate that our framework can both enhance models' performance and interpretability, we conducted extensive comparative experiments (Table \ref{table:main table}). The results yield the following findings:

\textbf{First}, \texttt{IBG} outperforms other models in terms of both performance and interpretability. Regarding performance, our framework performs better than the strong models like RGAT and DualGCN, as well as the latest ChatGPT model. As for interpretability, \texttt{IBG} also better captures meaningful words compared to classical gradient-based strategies and model-agnostic methods like IEGA. This indicates that \texttt{IBG} can effectively capture the sentiment-related low-dimensional features.

\textbf{Second}, \texttt{IBG} is capable of further enhancing both the performance and interpretability based on the existing models. 
We can see that our \texttt{IBG} results in a 1-2 points improvement in accuracy on each dataset. Taking BERT-SPC as an example, after integrating our framework, the Macro-F1 is improved by more than 10 points on the Lap14, Res15 and Res16 datasets. 
Even though Dual-GCN already performs well, Dual-GCN$_\texttt{IBG}$ still achieves an approximately 1\% increase in accuracy and 5\% improvement in F1 score over all datasets.
We also notice that our framework exhibits a higher improvement in Macro-F1. This indicates that the representations obtained by information bottleneck demonstrate higher sensitivity to different polarities, including neutral sentiment.
On the other hand, the effect of \texttt{IBG} on enhancing model interpretability is also significant. 
The models exhibit substantial increases in both AOPC and Ph-Acc after the incorporation of \texttt{IBG}. 
Compared to IEGA, the AOPC of BERT-SPC$_\texttt{IBG}$ surpasses BERT-SPC$_\texttt{IEGA}$ by 17.22\%, 9.02\% and 9.91\% on the Res14, Res15 and Res16 datasets, respectively.
These results further verify our \texttt{IBG}'s ability to better capture keywords like opinion words through learning the intrinsic dimension.

\textbf{Third}, \texttt{IBG} offers a better explanation in comparison to traditional gradient-based model interpretation strategies (Simple Gradient, Smooth Gradient and Integrated Gradient). 
BERT-SPC with \texttt{IBG} demonstrates more than twofold improvement in AOPC on the Res14 and Res16 datasets. 
Ph-Acc shows a remarkable 40.00\% improvement on Res14 compared with the Integrated Gradient method. Similarly, DualGCN-BERT$_\texttt{IBG}$ exhibits noticeable enhancements in AOPC compared to these three interpretation strategies, with the greatest improvements of 12.54\%, 15.53\%, 12.22\% and 7.23\% on the four datasets. 
However, the improvement in Ph-Acc on DualGCN-BERT is less pronounced. This can be attributed to the fact that DualGCN leverages GCN to capture the dependency relationships among words. Consequently, masking excessive words within a sentence can disrupt the integrity of the graph's structure, thus impacting the final prediction.
In summary, the incorporation of our framework is capable of enhancing models' ability to focus on contextually implicit sentiment information in words through gradient analysis, thereby improving its explanation.

\subsection{Ablation Studies}
We also conduct ablation experiments (Table \ref{table:ablation study}) to validate our framework from two perspectives: 
%
First, removing the iBiL (w/o iBiL) reduces performance, which indicates that compressing the model into a low-dimensional space for learning intrinsic dimension does benefit the ABSA task.
Second, the information bottleneck module has a significant impact on the model's final performance and interpretability. After removing the information bottleneck structure (w/o IB), the model's accuracy and F1-score both decline, even with a decrease of 4.07\% and 11.53\% in Ph-Acc on the Lap14 and Res14 datasets, respectively. This indicates that simply compressing embeddings to a lower dimensionality introduces noise, which affects the model's classification ability. Introducing an information bottleneck, on the other hand, does enable the model to forget irrelevant information and retain important features for sentiment classification.

\subsection{Further Analysis}
\begin{figure}[!t]
\centering
\subfigure[Lap14] {\includegraphics[scale=0.20]{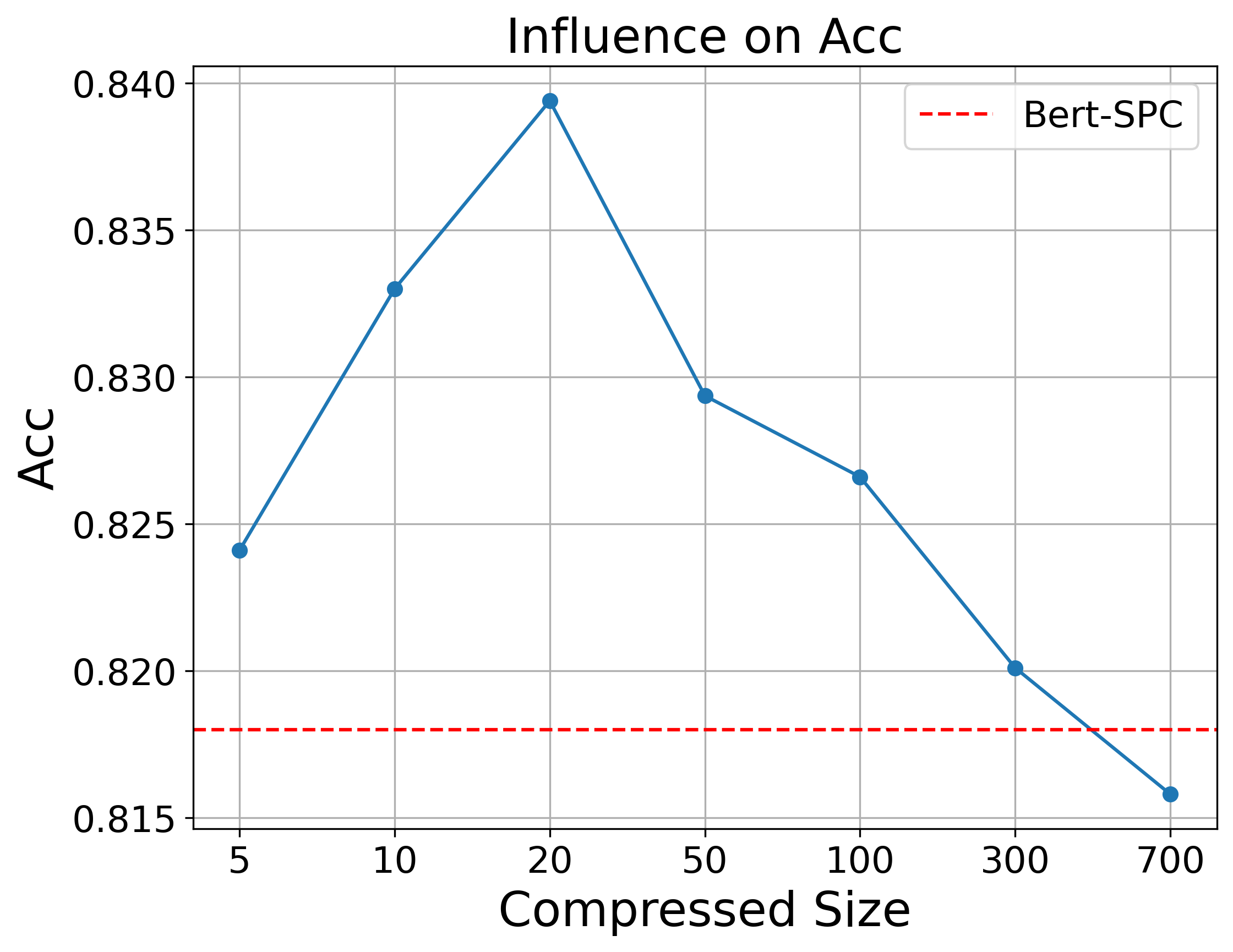}}
\subfigure[Lap14] {\includegraphics[scale=0.20]{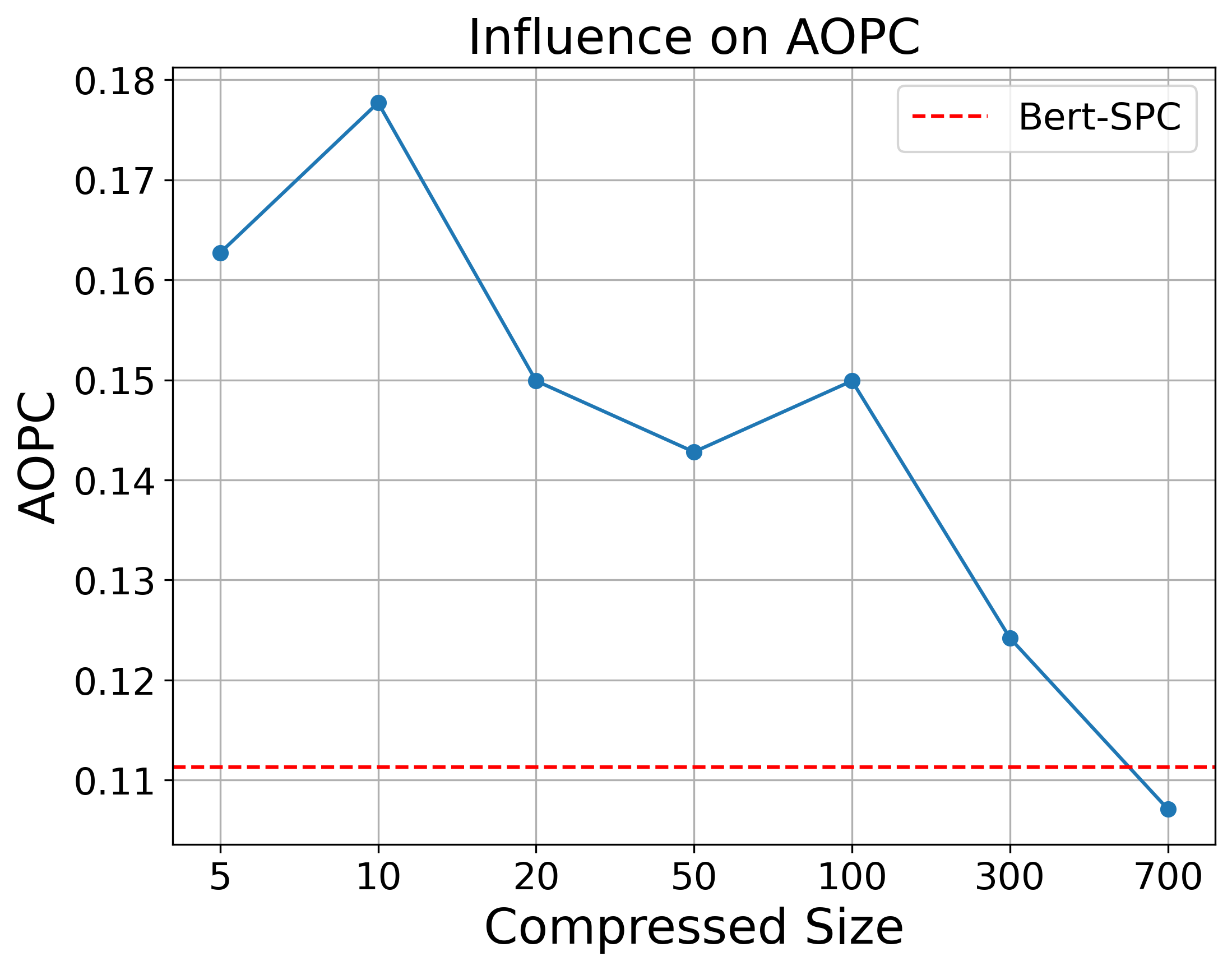}}
\subfigure[Res16] {\includegraphics[scale=0.20]{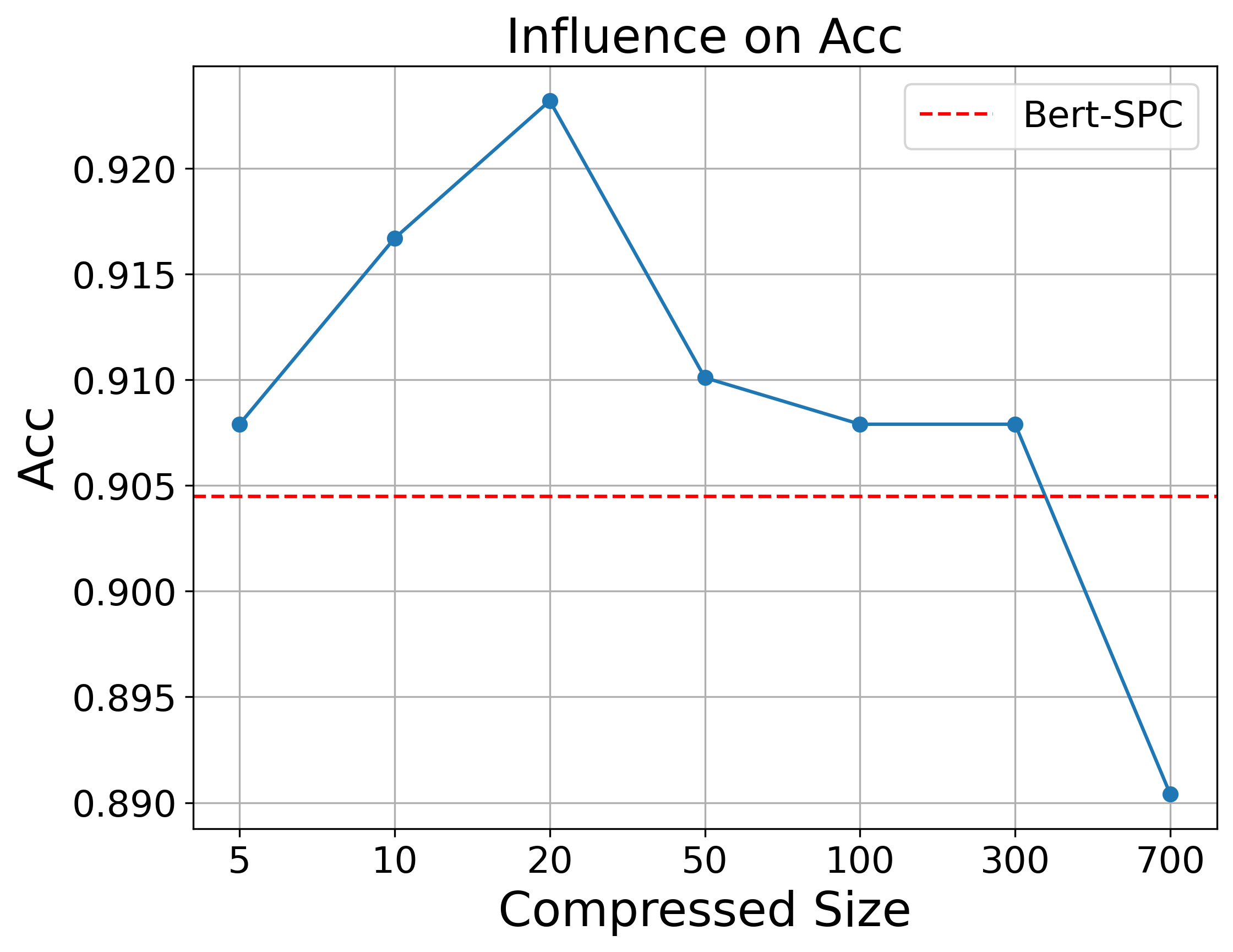}}
\subfigure[Res16] {\includegraphics[scale=0.20]{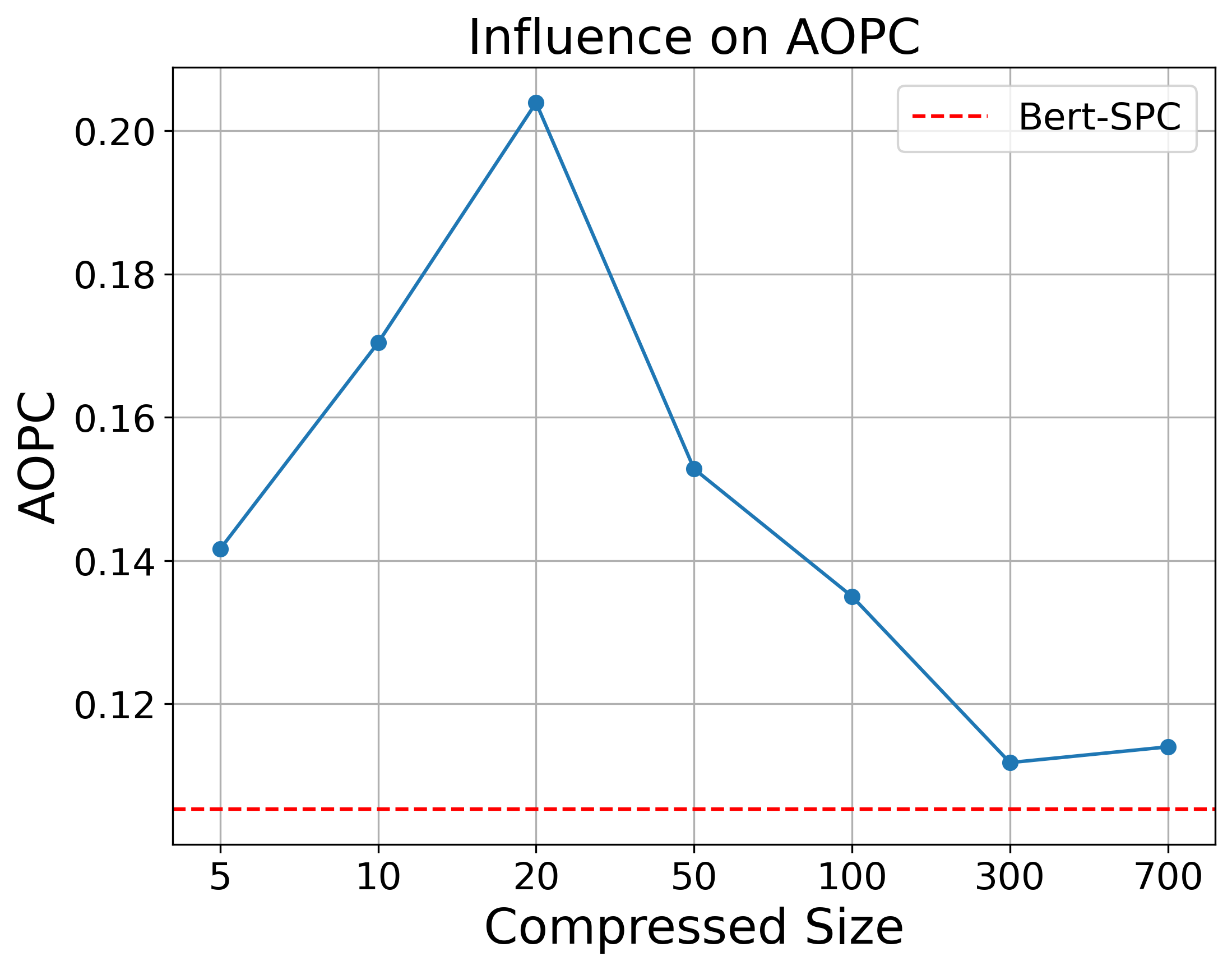}}
\caption{The influence of compressed size $\mathrm{Low}$.}
  \label{fig:The Influence of Compressed Size}
\end{figure}

\paragraph{The Influence of Compressed Size $\mathrm{Low}$.}
We further explore the variation in model performance and interpretability when compressing the pre-trained BERT embeddings into different dimensions. Two key findings can be deduced from Figure \ref{fig:The Influence of Compressed Size}:
First, with the continuous increase in dimensions, both models' performance and interpretability exhibit a trend of initially increasing and then decreasing.
Second, the optimal state of the model consistently reaches at 10 or 20 dimensions and the performance is even better than the original model without compressing. This suggests that our framework can effectively capture essential information in models' embeddings and eliminate redundant information.

\begin{figure}[!t]
\centering
\subfigure[AOPC] {\includegraphics[scale=0.15]{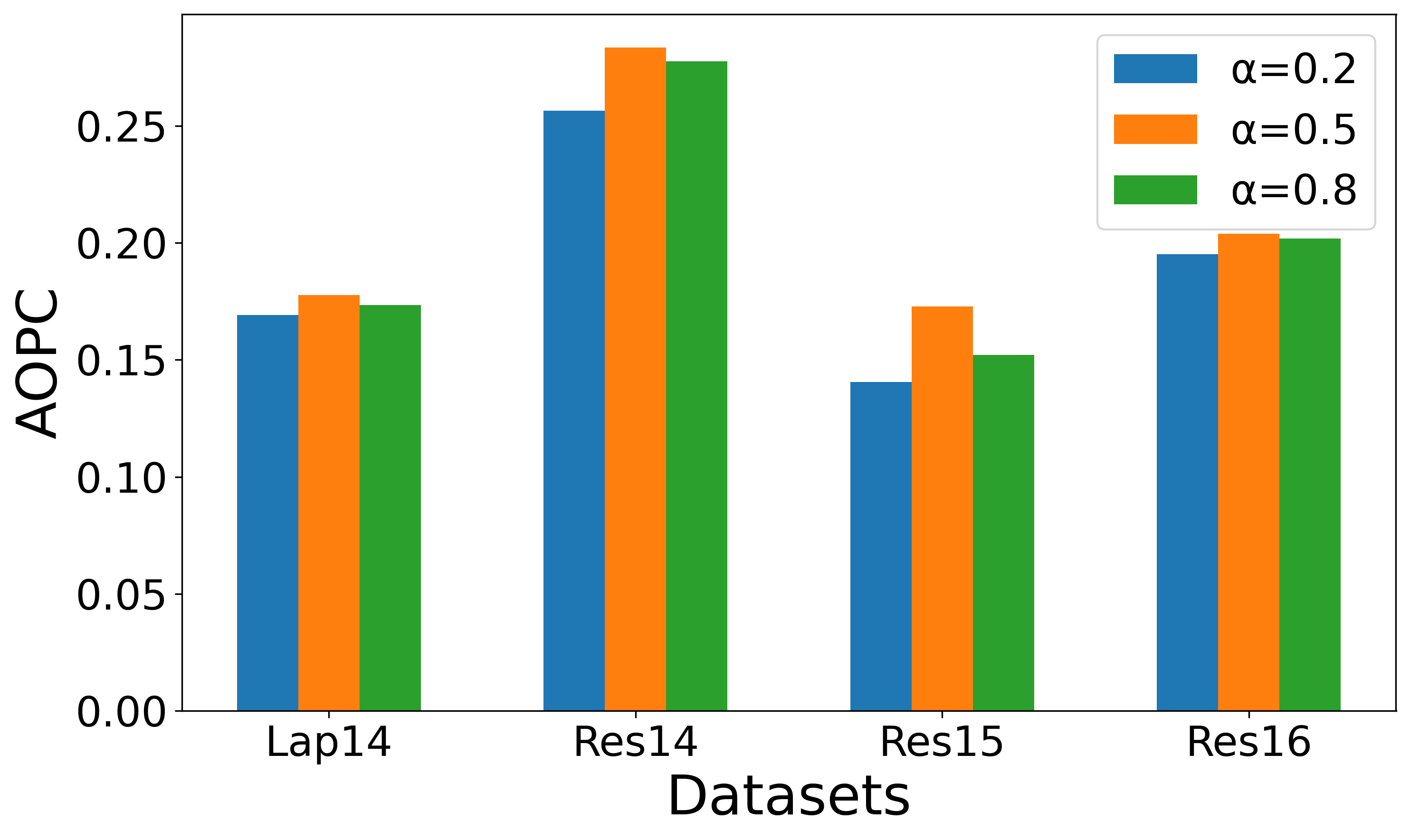}}
\subfigure[Ph-Acc] {\includegraphics[scale=0.15]{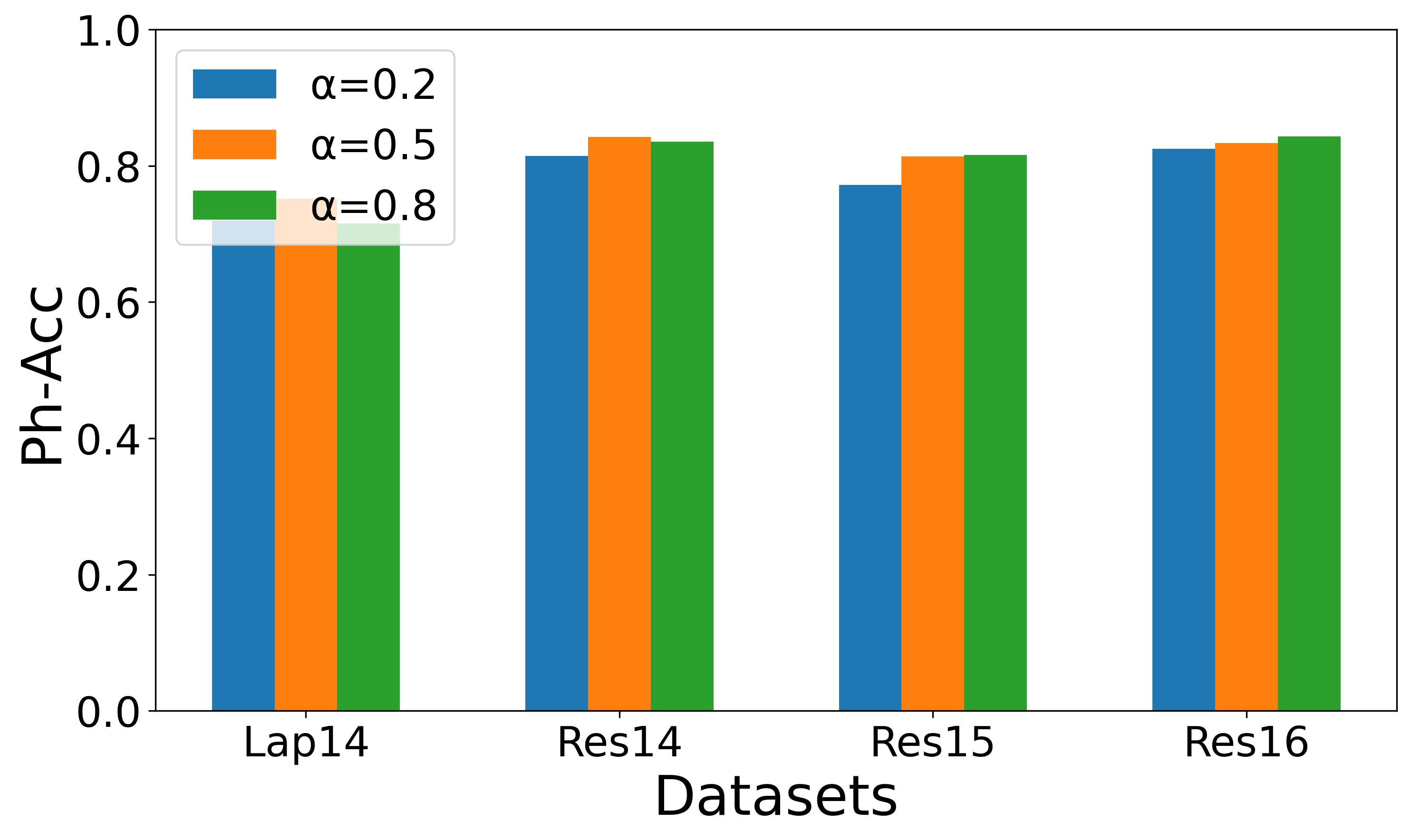}}
\caption{The influence of $\alpha$ on four datasets}
  \label{fig:influence of alpha}
\end{figure}

\paragraph{The Influence of $\alpha$.}
We conduct experiments to observe the impact of varying $\alpha$ values in Equation \ref{alpha equation}, which are used to balance the weights between the important scores obtained from intrinsic features and the original word embeddings (Figure \ref{fig:influence of alpha}). 
We can see that both the intrinsic representations and the original word embeddings are useful for the model's explanation. Relatively, assigning a higher weight to intrinsic dimension results in slightly higher AOPC and Ph-Acc compared with assigning a higher weight to word embeddings. It shows intrinsic vectors successfully learn the aspect-aware sentiment features, which is important for ABSA.

\section{Related Work}

\subsection{Intrinsic Dimension}
\citet{DBLP:conf/icml/OymakLS21} showed that trained neural networks often exhibit a low-rank property, giving rise to the concept of ``intrinsic dimension" \cite{li2018measuring}, which represents the minimum subspace dimension capable of encoding effective information or solving a problem. 
Following this concept, \citet{pope2020intrinsic} applied dimension estimation techniques to high-dimensional image data and discovered that natural image datasets have low intrinsic dimension. 
Significantly, \citet{aghajanyan2021intrinsic} first introduced the intrinsic dimension into the field of natural language processing, revealing that the size of the large pre-trained language models' intrinsic dimension is much smaller compared to that of their whole parameters.
After this, many studies started to incorporate low-rank structures into large models, achieving parameter-efficient learning by only tuning a small subspace based on the intrinsic dimension \citet{qin2021exploring,sun2022black,sun2022bbtv2,houlsby2019parameter,wang2022adamix,hu2021lora}.
%
Differing from predecessors' attempts to compress or tune models using intrinsic dimension, our work learns the intrinsic dimension of the large language models via information bottleneck, exploring how to better interpret the model's predictions in ABSA.

\subsection{Gradient-based Explanation Algorithms}
Literature on explanation and attribution methods has grown in the last few years, with a few broad categories
of approaches: perturbing the input \cite{fong2019understanding,ribeiro2016should}; utilizing gradient \cite{baehrens2010explain,binder2016layer,selvaraju2017grad}; visualizing intermediate layers \cite{zeiler2014visualizing}. 
Our work extends and improves upon the gradient-based method \cite{DBLP:journals/corr/SimonyanVZ13}, a popular technique applicable to many different types of models.
Several works were proposed to improve the original gradient-based methods, such as SmoothGrad \cite{DBLP:journals/corr/SmilkovTKVW17} and Integrated Gradient \cite{sundararajan2017axiomatic}.
Different from them, we optimize to compute gradient scores based on an intrinsic space, enabling a more effective model interpretation method.









\subsection{Information Bottleneck}
A series of studies motivated us to utilize IB \cite{DBLP:conf/emnlp/LiE19,zhou2021attending,zhou2022multi} to improve the explanations of gradient-based explanation methods. 
\citet{DBLP:conf/emnlp/LiE19} compressed the pre-trained embedding (e.g., BERT, ELMO), remaining only the information that helps a discriminative parser through variational IB.
\citet{zhmoginov2019information} utilized the IB approach to discover the salient region.
Some works \cite{jiang2020inserting,chen2018learning,guan2019towards,DBLP:conf/iclr/SchulzSTL20,bang2019explaining} proposed to identify vital features or attributions via IB. 
Moreover, \citet{chen2020learning} designed a variational mask strategy to delete the useless words in the text. In this paper, we utilize IB to learn the intrinsic space to improve the models' explanation.

\subsection{Aspect-based Sentiment Analysis}
Aspect-based Sentiment Analysis (ABSA) involves the extraction of aspect terms and opinion words from a sentence and the prediction of sentiment polarity \cite{zhang2022survey}. 
In our study, we focus on the subtask Aspect-based Sentiment Classification (ABSC), which entails predicting sentiment labels for a given sentence and its associated aspect. 
To consider the complex contextual relationships in sentences, some ABSC research combined attention mechanisms with large pre-trained language models, such as BERT-SPC \cite{kenton2019bert}, AEN-BERT \cite{song2019attentional} and LCF-BERT \cite{zeng2019lcf}. 
There is another trend of combining dependency trees and Graph Convolutional Networks (GCNs), exploiting syntax information explicitly, like RGAT \cite{wang2020relational}, DualGCN \cite{li2021dual} and SSEGCN \cite{zhang2022ssegcn}.
However, even when employing methods such as dependency trees to align aspect terms with their corresponding opinion words, in practice, we still observe that the model may focus on the wrong aspect, especially in sentences containing multiple aspects. Therefore, research on Explainable Aspect-based Sentiment Analysis is essential.
\citet{cheng2023tell} leveraged annotated opinion words to force the model to pay greater attention to these words in terms of gradients, enhancing the models' interpretability, but the cost of annotation is high. 
In this paper, we propose a model-agnostic framework to enhance both performance and interpretability without additional labels.

\section{Conclusions and Further Work}
This paper conducts preliminary experiments, demonstrating the uneven importance of word embedding dimensions in ABSA. However, the current gradient-based explanation methods do not take this difference into account.
Thus, we propose an Information Bottleneck-based Gradient (\texttt{IBG}) explanation framework for ABSA, leveraging the information bottleneck principle to compel the model to learn intrinsic information. 
By integrating our framework with the latest models, we conduct extensive comparative experiments, confirming that our proposed \texttt{IBG} framework significantly enhances both the performance and interpretability of the original models. 
Through ablation experiments, we demonstrate the beneficial impact of the information bottleneck structure and the attempt to map the embedding layer to a low-dimensional intrinsic space. 
Future research will explore applying \texttt{IBG} to large-scale language models (e.g., LLaMA) and other NLP tasks.

\section*{Acknowledge}
The authors wish to thank the reviewers for their helpful comments and suggestions. This research is funded by the National Key Research and Development Program of China (No.2021ZD0114002), the National Natural Science Foundation of China (No.62307028 and No.62377013), the Science and Technology Commission of Shanghai Municipality Grant (No.22511105901, No.21511100402 and No.21511100302), and Shanghai Science and Technology Innovation Action Plan (No.23ZR1441800 and No.23YF1426100). 

\section*{References}
\bibliographystyle{lrec_natbib}
\bibliography{lrec-coling2024-example}




\end{document}